\documentclass[10pt,twocolumn,letterpaper]{article}

\usepackage{cvpr}
\usepackage{times}
\usepackage{epsfig}
\usepackage{graphicx}
\usepackage{amsmath}
\usepackage{amssymb}
\usepackage{color}
\usepackage{indentfirst}
\usepackage{bm}
\usepackage{gensymb}
\definecolor{myRed}{rgb}{0.8, 0.2, 0.2}
\definecolor{myBlue}{rgb}{0.2,0.2,0.8}

\newcommand{\figref}[1]{Fig.~\ref{#1}}

\newcommand{\tabref}[1]{Tab.~\ref{#1}}

\usepackage[breaklinks=true,bookmarks=false]{}

\cvprfinalcopy 

\def\cvprPaperID{****} 

\begin{document}

\title{DeepLiDAR: Deep Surface Normal Guided Depth Prediction for Outdoor Scene from Sparse LiDAR Data and Single Color Image}

\author{
	Jiaxiong Qiu$^{1*}$ \quad  Zhaopeng Cui$^{2*}$ \quad  Yinda Zhang$^{3}$\thanks{indicates equal contributions.} \\  Xingdi Zhang$^{1}$ \quad Shuaicheng Liu$^{1,4}$\thanks{indicates corresponding author.} 
	\quad Bing Zeng$^{1}$ \quad Marc Pollefeys$^{2,5}$\\
	$^{1}$ UESTC \quad $^{2}$ETH Z\"{u}rich \quad  $^{3}$Google  \quad  $^{4}$Megvii Technology \quad $^{5}$Microsoft \vspace{-0.3em}
}

\maketitle

\begin{abstract}
In this paper, we propose a deep learning architecture that produces accurate dense depth for the outdoor scene from a single color image and a sparse depth.
Inspired by the indoor depth completion, our network estimates surface normals as the intermediate representation to produce dense depth, and can be trained end-to-end. 
With a modified encoder-decoder structure, our network effectively fuses the dense color image and the sparse LiDAR depth.
To address outdoor specific challenges, our network  predicts a confidence mask to handle mixed LiDAR signals near foreground boundaries due to occlusion, and combines estimates from the color image and surface normals with learned attention maps to improve the depth accuracy especially for distant areas.
Extensive experiments demonstrate that our model improves upon the state-of-the-art performance on KITTI depth completion benchmark. Ablation study shows the positive impact of each model components to the final performance, and comprehensive analysis shows that our model generalizes well to the input with higher sparsity or from indoor scenes.

\end{abstract}

\vspace{-0.5em}
\section{Introduction}
Measuring dense and accurate depth for outdoor environment is critically important for various applications, such as autonomous driving and unmanned aerial vehicles.
Most of the active depth sensing solutions for indoor environment fail due to strong interference of the passive illumination \cite{fanello2017ultrastereo,hyperdepth}, and 
stereo methods usually become less accurate for distant areas due to lower resolutions and smaller triangulation angles compared to the close areas \cite{Szeliski_book}.
As a result, LiDAR is the dominating reliable solution for the outdoor environment.
However, the high-end LiDAR is prohibitively expensive, and the commodity level devices suffer from the notorious low resolution \cite{lingfors2017comparing} which causes troubles for perception in middle or long range area.
Spatial and temporal fusion provide denser depth but either requires multiple devices or suffers from dynamic objects and latency.
An affordable solution for immediate access of the dense and accurate depth still does not exist.
\begin{figure}[t]
	\centering
	\includegraphics[width=\linewidth]{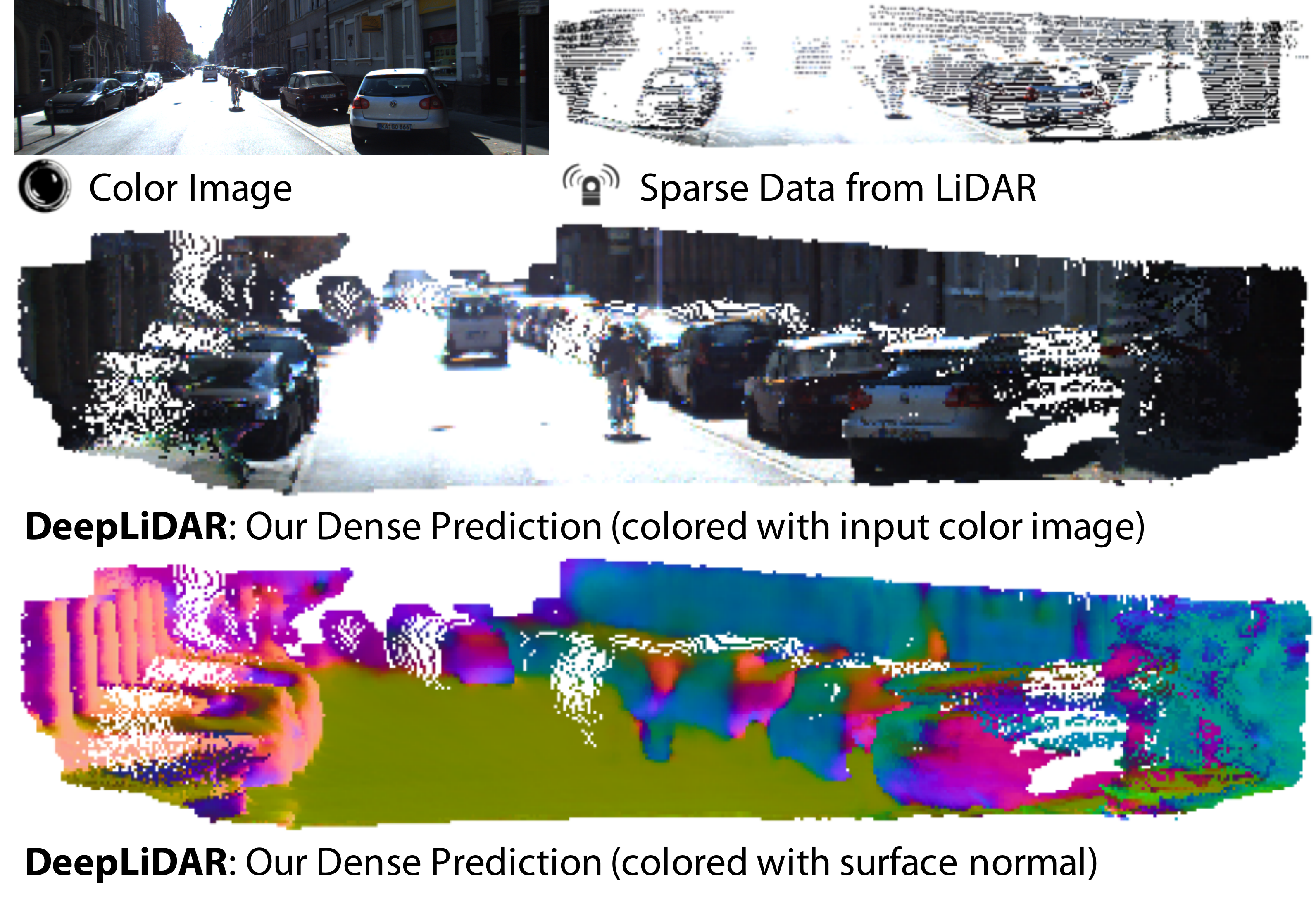}
	\vspace{-1.5em}
	\caption{{\bf Our system} takes as input a color image and a sparse depth image from the LiDAR (Row 1), and output a dense depth map (Row 2). Rather than directly producing the complete depth, our model estimates surface normals (Row 3) as the intermediate representation which is helpful to produce accurate depth. }
	\label{fig:teaser}
\end{figure}

One promising attempt is to take a sparse but accurate depth from a low-cost LiDAR and make it dense with the help of an aligned color image.
With the great success of deep learning, an obvious approach is to directly feed the sparse depth and color image into a neural network and regress for the dense depth. 
Unfortunately, such a black-box does not work equally well compared to interpretable models, where local depth affinity is learned from color image to interpolate the sparse signal.
For indoor scenes, Zhang \etal \cite{zhang2018deepdepth} estimated the surface normals as the intermediate representation and solved for depth via a separate optimization, which achieved superior results.
However, it is not well studied if the surface normal is a reasonable representation for the outdoor scene and how such system performs.

In this work, we propose an end-to-end deep learning system to produce dense depth from sparse LiDAR data and a color image taken from outdoor on-road scenes leveraging surface normal as the intermediate representation.
We find it non-trivial to make such a system work equally well as in the indoor environment, generally because of the following three challenges: 

\vspace{0.5pt}
\noindent\textbf{Data Fusion.}
How to combine the given sparse depth and dense color image is still an open problem.
One common manner is to concatenate them (usually with a binary mask indicating the pixel-wise availability of the LiDAR depth) directly as the network input (i.e. early fusion), in which the network has the best access to all sources of inputs starting from the encoder.
However, the result may produce artifacts near the boundaries of the missing values, or merely copy depth from where it is available but fail otherwise.
Inspired by the idea of leveraging intermediate affinity, we design an encoder-decoder architecture, namely deep completion unit (DCU), where separate encoders learn affinity from the color image and features from the sparse depth respectively, while the decoder learns to produce dense output.
The DCU falls in the style of late fusion architecture but different in that the feature from the sparse depth is summed into the decoder rather than ordinary concatenation.
The summation \cite{chen2017multi} favors the features on both sides in the same domain, and therefore encourages our decoder to learn features more related with depth in order to keep consistent with the feature from the sparse depth.
This also saves network parameters as well as inference memory.
Empirically, we find DCU benefits both the intermediate surface normal and the final depth estimation.


\begin{figure}[t]
	\centering
	\includegraphics[width=\linewidth]{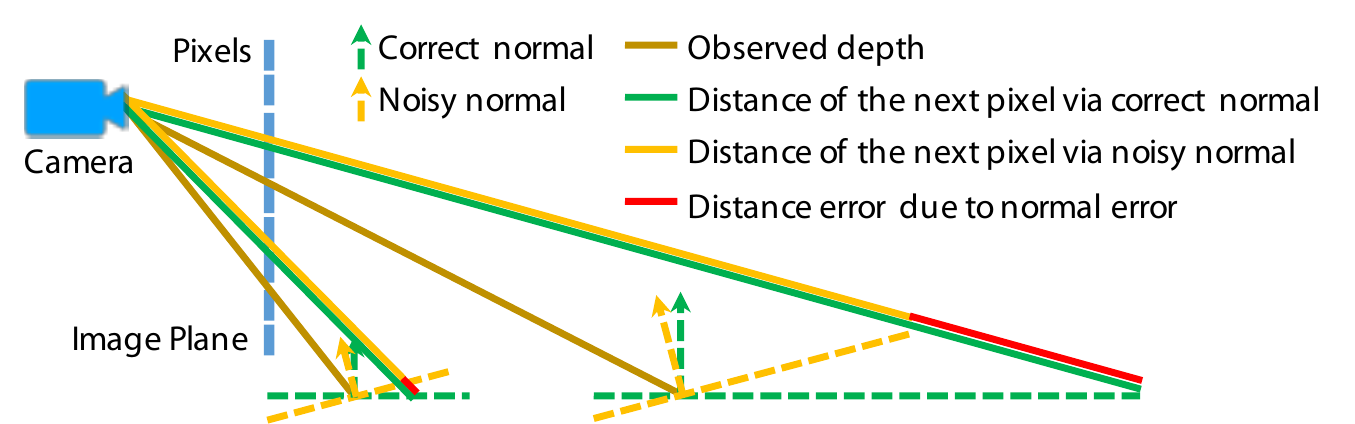}
	\caption{{\bf Sensitivity to noise.} Reconstructing depth from normal becomes more sensitive to the noise/error in the estimated normal when the distance goes up. We show two cases to estimate the depth of the neighboring pixel via correct (green) and noisy (yellow) normal. The further case results in much larger error (red) compared to the closer one even though the surface normal error is the same ($15\degree$) for two cases. }
	\label{fig:failure}
\end{figure}

\vspace{0.5pt}
\noindent\textbf{Sensitivity to Noise.}
Zhang \etal \cite{zhang2018deepdepth} demonstrated that surface normals of indoor scenes are easier to estimate than absolute depth and sufficient to complete the depth given incomplete signals. 
However, in outdoor scenes,
solving depth from normals does not work ubiquitously well especially for the distant area mainly due to the perspective geometry.
As shown in \figref{fig:failure},
the same surface normal error causes much larger distance error for the horizontal road surface in the far area compared to the close range area.
Having these areas hard to be solved from surface normals geometrically, we propose to learn them directly from the raw inputs.
Therefore, our model contains two pathways to estimate dense depth maps from the estimated surface normals and the color image respectively, which are then integrated via automatically learned attention maps.
In other words, the attention maps learn to collect better solution for each area from the pathway that is likely to perform better.

\vspace{2.0pt}
\noindent\textbf{Occlusion.}
As there is almost inevitably a small displacement between the RGB camera and the LiDAR sensor, different depth values are normally mixed with each other along the boundaries due to occlusion when warping LiDAR data to the color camera coordinate, especially for the regions close to the camera (\figref{fig:misalignemnt} (b)). 
Such mixture of depth confuses the model and causes blurry boundaries. 
Ideally, the model should downgrade the confidence of the sparse depth in these confusing area and learn to fill in using more reliable surroundings.
We propose to learn such a confidence mask automatically, which takes the place of the binary availability mask feeding into the surface normal pathway.
Even though without ground truth, our model self-supervisely learns this occlusion area containing overlapping sparse depth.

\begin{figure*}[t]
\vspace{-4mm}
	\centering
	\includegraphics[width=0.88\linewidth]{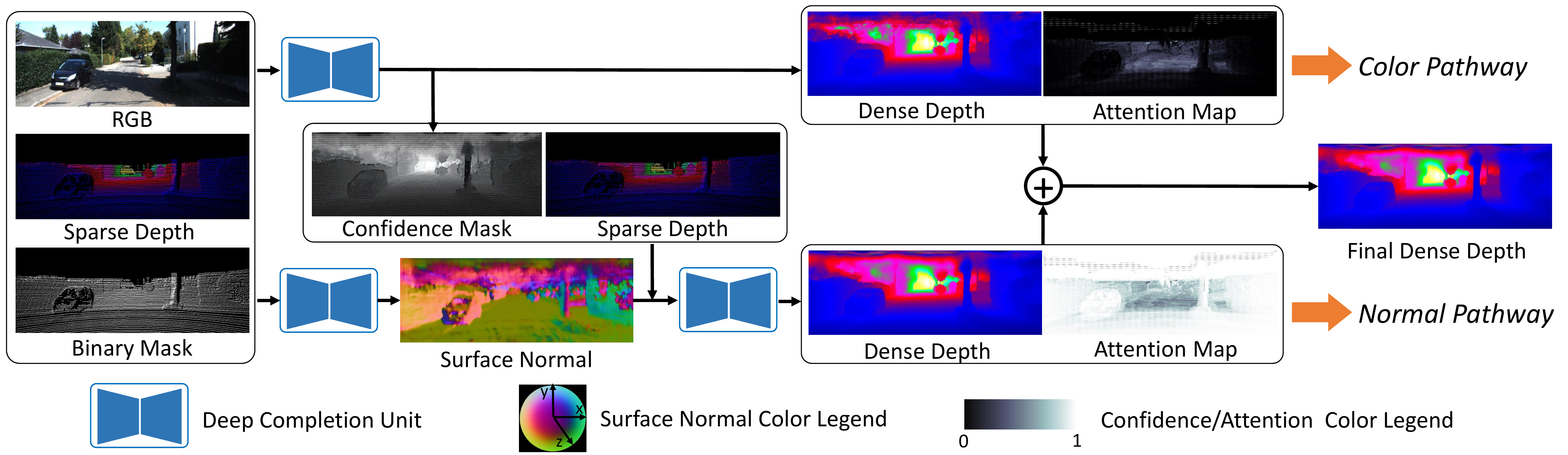}
	\vspace{-0.3em}
	\caption{{\bf The pipeline of our model.} Our model consists of two pathways. Both starting from a RGB image, a sparse depth, and a binary mask as the inputs, the surface normal pathway (lower half) produces a pixel-wise surface normal for the scene, which is further combined with the sparse input depth and a confidence mask estimated from the color pathway to produce a dense depth. The color pathway produces a dense depth too. The final dense depth output is the weighted sum of the depths from two pathways using the estimated attention maps.}
	\label{fig:pipeline}
	\vspace{-1.5em}
\end{figure*}

Our full pipeline is shown in  \figref{fig:pipeline}. The contributions of this work are as follows. Firstly, we propose an end-to-end neural network architecture that produces dense depth from a sparse LiDAR depth and a color image using the surface normal as the intermediate representation, and demonstrate that surface normal is also a good local depth representation for the outdoor scene.
Secondly, we propose a modified encoder-decoder structure to effectively fuse the sparse depth and the dense color image.
Thirdly, we investigate the challenges for outdoor scenarios, and design the network to automatically learn a confidence mask for occlusion handling, and attention maps for the integration of depth estimates from both the color image and normals.
Lastly, our experiment shows that out model significantly outperforms the state-of-the-art on benchmarks and generalizes well to input sparsity and indoor scenes.

\section{Related Work}
\noindent\textbf{Depth prediction from sparse samples.}
Producing dense depth from sparse inputs starts to draw attention when accurate but low-res depth sensors, such as low-cost LiDAR and one-line laser sensors, become widely available.
Some methods produced dense depth or disparity via wavelet analysis \cite{hawe2011dense,liu2015depth}.
Recently, deep learning based approaches were proposed and achieved promising results. 
Uhrig \etal \cite{uhrig2017sparsity} proposed sparsity invariant CNNs to deal with the variant input depth sparsity.
Ma \etal \cite{Ma2018SparseToDense} proposed to feed the concatenation of the sparse depth and the color image into an encoder-decoder deep network, and further extended with self-supervised learning \cite{ma2018self}.  
Jaritz \etal \cite{jaritz2018sparse} combined semantic segmentation to improve the depth completion.
Cheng \etal \cite{cheng2018depth} learned an affinity matrix to guide the depth interpolation through a recurrent neural network.
Compared to these work, our model is more physically driven and explicitly exploits surface normals as the intermediate representation.


\vspace{0.5pt}
\noindent\textbf{Depth refinement for indoor environment.}
In the indoor environment, the quality of the depth from commodity RGB-D sensors is not ideal due to the limitation of the sensing technologies \cite{bhandari14b,naik15,fanello2017ultrastereo}.
A lot of works have been proposed to improve the depth using an aligned high-resolution color image.
One family of approach is depth super-resolution that targets improving the resolution of the depth image \cite{mac2012patch,shabaninia2017high,han2013high,yu2013shading,lu2015sparse,kiechle2013joint,mahmoudi2012sparse,tosic2014learning}. 
These methods assume a low-resolution but dense depth map without missing signal.
The other family of methods is color image guided depth inpainting, which potentially handles large missing area with arbitrary shape.
Traditional methods use color as the guidance to compute local affinity or discontinuity \cite{herrera2013depth,gong2013guided,silberman2012indoor,barron2016fast,ferstl2013image,zhang2017probability,zuo2016explicit,barron2013intrinsic}.
Even though deep learning has been widely used in image inpainting \cite{van2016conditional,pathak2016context,liu2018image,yu2018free}, extension of these networks to color guided depth inpainting is not well studied.
Zhang \etal \cite{zhang2018deepdepth} proposed to estimate surface normals and solve for depth via a global optimization. 
However, it is still unclear if normals, as the intermediate representation for depth, still work for the outdoor scenes.

\vspace{0.5pt}
\noindent\textbf{Depth estimation from a single RGB image.}
There are a lot of works estimating depth from only a single color image.
Early methods mainly relied on the hand-crafted features and probabilistic graphical models \cite{saxena2006learning, karsch2012depth, karsch2014depth, konrad20122d, liu2014discrete}.  
With the development of deep learning, many methods \cite{eigen2014depth, laina2016deeper, roy2016monocular, liu2016learning} based on deep neural networks have been proposed for the single-view depth estimation due to the strong feature representation of deep networks. For example, Eigen \etal \cite{eigen2014depth} proposed a multi-scale convolutional network to predict depth from coarse to fine. Laina \etal \cite{laina2016deeper} proposed a single-scale but deeper fully convolutional architecture. Liu \etal \cite{liu2016learning} combined the strength of deep CNN and continuous CRF in a unified CNN framework. 
There are also some methods \cite{eigen2015predicting, li2015depth, chen2017surface, qi2018geonet} which exploit surface normals during the depth estimation. Eigen \etal \cite{eigen2015predicting} and Li \etal \cite{li2015depth} proposed architectures to predict depth or normals but independently. Chen \etal \cite{chen2017surface} used sparse surface annotation as supervision for depth estimation but not intermediate representation. Qi \etal \cite{qi2018geonet} jointly predicted depth and surface normal based on two-stream CNNs and focused on indoor scenes.
Most recently, some unsupervised methods \cite{zhou2017unsupervised, godard2017unsupervised, yang2018lego} were also proposed.
Even though these methods produced plausible depth estimation from a single color image, they do not handle sparse depth as an additional input and are not suitable to recover high-quality depth.
Moreover, our method is the first to use surface normals as the intermediate representation for the outdoor depth completion.

\vspace{-0.2em}
\section{Method}
\vspace{-0.2em}
Our model is an end-to-end deep learning framework that takes an RGB image and a sparse depth image reprojected from LiDAR as inputs and produce a dense depth image.
As illustrated in \figref{fig:pipeline}, the whole network mainly consists of two pathways: the color pathway and surface normal pathway.
The color pathway takes as input the color image and the sparse depth to output a complete depth.
The surface normal pathway first predicts a surface normal image from the input color image and sparse depth, which is then combined together with the sparse depth and a confidence mask learned from the color pathway to produce a complete depth.
Each of these two pathways are implemented with a stack of deep completion units (DCU), 
and the depths from two pathways are then integrated by a learned weighted sum to produce the final complete depth.

\vspace{-0.2em}
\subsection{Deep Completion Unit}
\vspace{-0.2em}
Zhang \etal \cite{zhang2018deepdepth} proposes to remove the incomplete depth from the input when predicting either depth or surface normal in order to get rid of the local optima.
However, since the sparse depth is strongly correlated with the dense depth and surface normals, it is certainly non-optimal if the network has no chance to learn from it.
Inspired by traditional color image guided inpainting \cite{tomasi1998bilateral,he2010guided,yu2018free}, 
we propose a network architecture to have the encoder to learn the local affinity from color image or surface normals, which is then leveraged by the decoder to conduct interpolation with the features generated from the input sparse depth through another encoder.

\begin{figure}[t]
	\centering
	\includegraphics[width=0.98\linewidth]{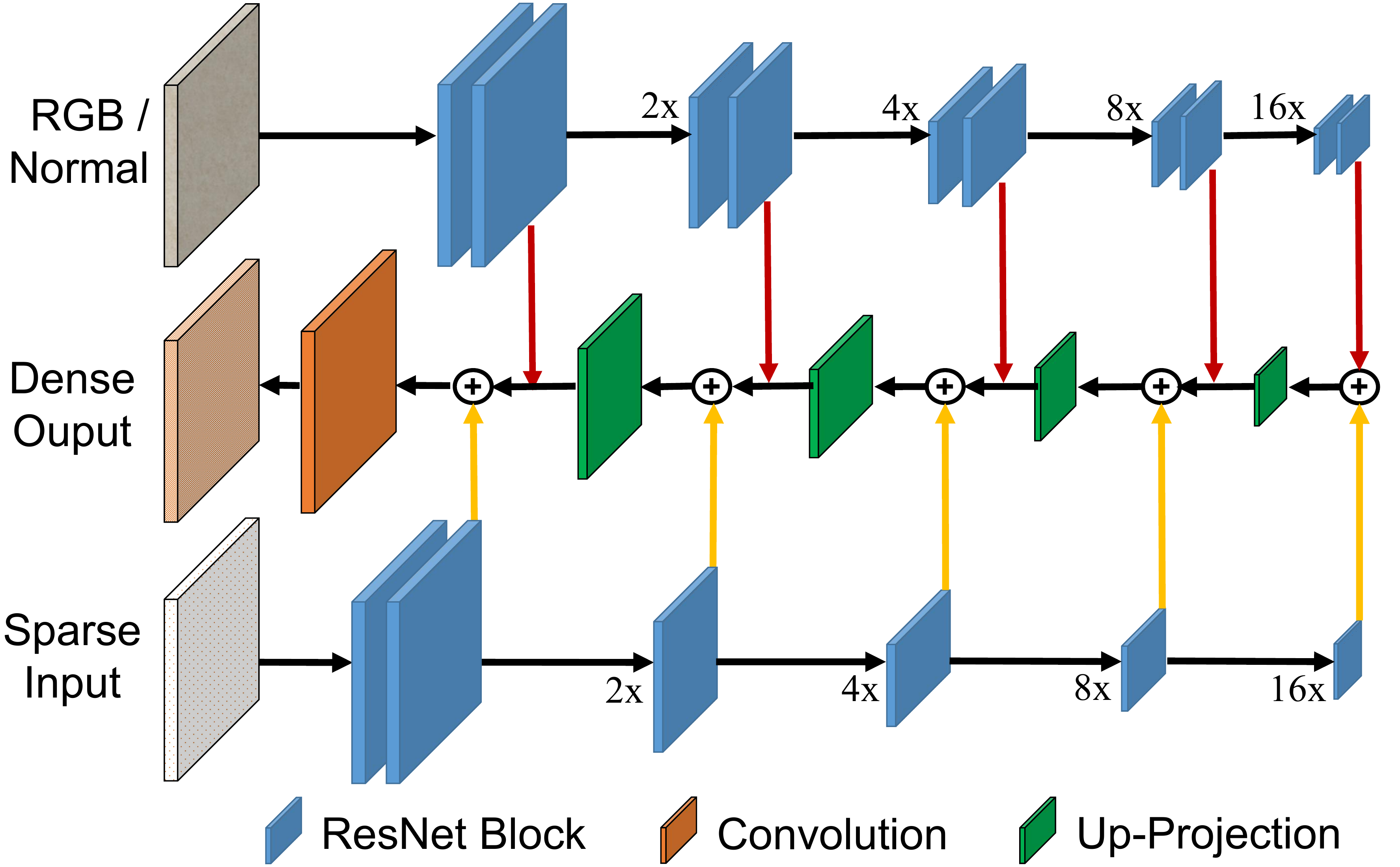}
	\vspace{-0.5em}
	\caption{{\bf Detailed architecture of deep completion unit.} Our deep completion unit takes the late fusion strategy, where features from the RGB/normal and sparse depth are combined only in the decoder. Different from \cite{jaritz2018sparse}, we sum the features from both side at each resolution throughout the decoder.}
	\label{fig:DFU}
\end{figure}

The details of our deep completion unit is shown in \figref{fig:DFU}. 
Both encoders for RGB/normal and sparse depth consist of a series of ResNet blocks followed by convolution with stride to downsize the feature resolution eventually to $1/16$ of the input.
The decoder consists of four up-projection units as introduced in \cite{laina2016deeper} to gradually increase the feature resolutions and to integrate features from both encoders to produce dense output.
Since the input sparse depth is strongly related with the decoder output, e.g., surface normal or depth, features from the sparse depth should contribute more in the decoder.
As such, we concatenate the features from the RGB/normal but sum the features from the sparse depth onto the features in decoder.
As the summation favors the features on both sides in the same domain \cite{chen2017multi}, the decoder is encouraged to learn features more related to depth in order to keep consistent with the feature from the sparse depth.
As shown in \figref{fig:pipeline}, we use the DCU to predict either surface normal or depth with the same input but trained with the target ground truth.

\vspace{-0.2em}
\subsection{Attention Based Integration}
\vspace{-0.2em}
Recovering depth from surface normals does not work ubiquitously well everywhere, and might be sensitive to normal noise in some areas.
We propose to generate depth for these areas leveraging priors from the color image rather than geometry from the estimated surface normal.
Therefore, our model consists of two pathways in parallel to predict dense depth from the input color image and estimated surface normals respectively.
Both pathways also take the sparse depth as input.
The final dense depth should be an integration of these two estimated depths, where comparatively more accurate depth measurements are chosen from the right one. 

We use an attention mechanism to integrate the depths recovered from two pathways, where the combination of two depths is not fixed but depends on the current context.
In particular, we first predict a score map for each of pathways using the last feature map before the output through three convolutions with ReLU.
The two score maps from two pathways are then fed into a softmax layer, and converted into a combination weight.
The final dense depth output is then calculated as 
\vspace{-0.4em}
\begin{equation}
\Hat{D} = w_c \cdot \Hat{D_c} + w_n \cdot \Hat{D_n},
\label{dp}
\vspace{-0.4em}
\end{equation}
where $\Hat{D_c}$ and $\Hat{D_n}$ are depths from color and surface normal pathway, and $w_c$ and $w_n$ are the learned combination weights respectively.
As it can be seen in \figref{fig:kitti_valid_quality}, the learned $w_c$ and $w_n$ target on the strong part of their corresponding depth output effectively.


\begin{figure}[t]
	\centering
	\begin{tabular}{*{3}{c@{\hspace{6px}}}}
		\multicolumn{3}{c}{\includegraphics[width=0.35\textwidth]{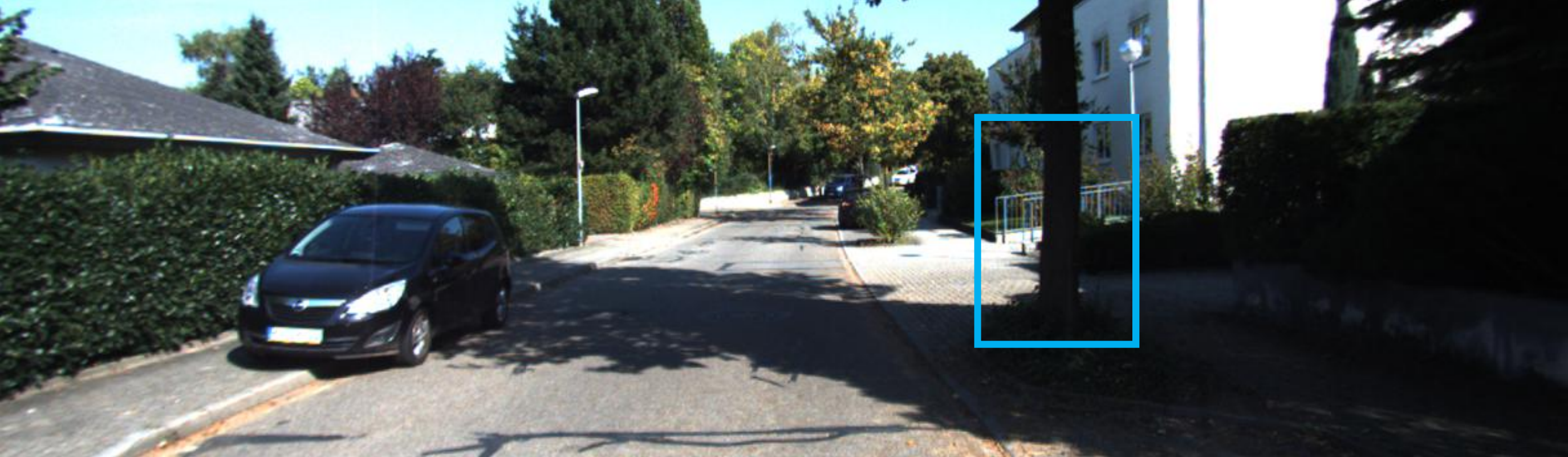}} \\
		\multicolumn{3}{c}{\small{(a) RGB Image}}\\
		\includegraphics[width=0.13\textwidth]{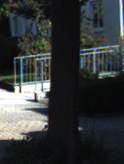} &  
		\includegraphics[width=0.13\textwidth]{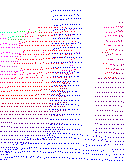}&
		\includegraphics[width=0.13\textwidth]{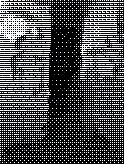}\\
		\small{(b) Zoom-in view} & \small{(c) Warped depth} & \small{(d) Confidence} \\
	\end{tabular}
	\caption{{\bf Occlusion and learned confidence.} (b) shows a zoom-in view of the region marked by the blue box in (a). Due to the displacement between the RGB camera and LiDAR, aligning sparse depth to the color image causes foreground/background depth mixed in the occluded area, like the tree trunk in (c). Our network learns a confidence mask (d) that successfully down-weight the confusing mixed area.}
	\label{fig:misalignemnt}
\end{figure}

\subsection{Confidence Prediction}
\vspace{-0.1em}
As mentioned before and shown in \figref{fig:misalignemnt}, there are ambiguous areas with mixture of foreground and background depth signals due to the displacement between the LiDAR sensor and the color camera.
This is usually caused by occlusion, which happens more frequently along the object boundaries in close range.
Ideally, we should find these confusing areas and resolve the ambiguity, which however is even more challenging as this requires an accurate 3D geometry estimation near the depth discontinuities.
On the contrary, we ask the network to automatically learn a confidence mask to indicate the reliability of the input sparse depth.
We replace the simple binary mask, 
which is an input hard confidence, 
with the learned confidence mask ($m_d$) from the color pathway. As shown in \figref{fig:misalignemnt}, even though without ground truth of such masks, the model could successfully learn the occlusion area with overlapping sparse depth values (e.g., low weights for tree trunk).

\vspace{-0.2em}
\subsection{Loss Function}
\vspace{-0.2em}
The loss function of the overall system is defined as:
\vspace{-0.3em}
\begin{equation}
L = \lambda_1 L_d(\Hat{D_n}) + \lambda_2 L_d(\Hat{D_c}) + \lambda_3 L_d(\Hat{D}) + \lambda_4 L_n(N)
\label{loss_depth}
\vspace{-0.3em}
\end{equation}
where $L_d$ defines the loss on the estimated depth, and $L_n$ defines the loss on the estimated surface normal.
We use cosine loss following \cite{zhang2017physically} for $L_n$.
For $L_d$, we use $L_2$ loss 
on the estimated depth and a cosine loss on the normal converted from the depth.
$\lambda_{1,2,3,4}$ adjusts the weights between terms of the loss function.
We adopt a multi-stage training schema for stable convergence. 
We first set $\lambda_4=1$ and all the other weights to zero to only pre-train the surface normal estimation. We then set $\lambda_1=0.3, \lambda_2=0.3, \lambda_3=0.0, \lambda_4=0.1$ to further train the color and surface normal pathways. In the end, we set $\lambda_1=0.3, \lambda_2=0.3, \lambda_3=0.5, \lambda_4=0.1$ to train the whole system end-to-end.
For all the training setting, we use Adam as the optimizer with a starting learning rate of $0.001$, $\beta_1$ = $0.9$ and $\beta_2$ = $0.999$. The learning rate is descended to half every 5 epochs.

\vspace{-0.2em}
\subsection{Training Data}
\vspace{-0.2em}
Due to the lack of the ground-truth normal in the real datasets, we generate a synthetic dataset using an open urban driving simulator Carla~\cite{Dosovitskiy17}. 
We render 50K training samples including RGB image, sparse depth map, dense depth map, and surface normal image, and the examples 
are shown in our supplementary materials. 
For the real data, we use the KITTI depth completion benchmark dataset for finetuning and evaluation. The complete surface normal ground truth for the KITTI dataset is computed from the ground-truth dense depth map by local plane fitting \cite{silberman2012indoor}.

\section{Experiments}
\vspace{-0.2em}


We perform extensive experiments to verify the effectiveness of our model, including comparison to related work and ablation study.
Since one of the major applications of our model is on car-held LiDAR devices, most of the experiments are done on KITTI depth completion benchmark \cite{uhrig2017sparsity}.
Nevertheless, we also run our model in indoor environment to verify the generalization capability.

\vspace{-0.2em}
\subsection{Comparison to State-of-the-art}
\begin{figure*}[h]
\vspace{-4mm}
	\centering
	\begin{tabular}{*{3}{c@{\hspace{3.5px}}}}
		\multicolumn{3}{c}{\includegraphics[width=0.99\textwidth]{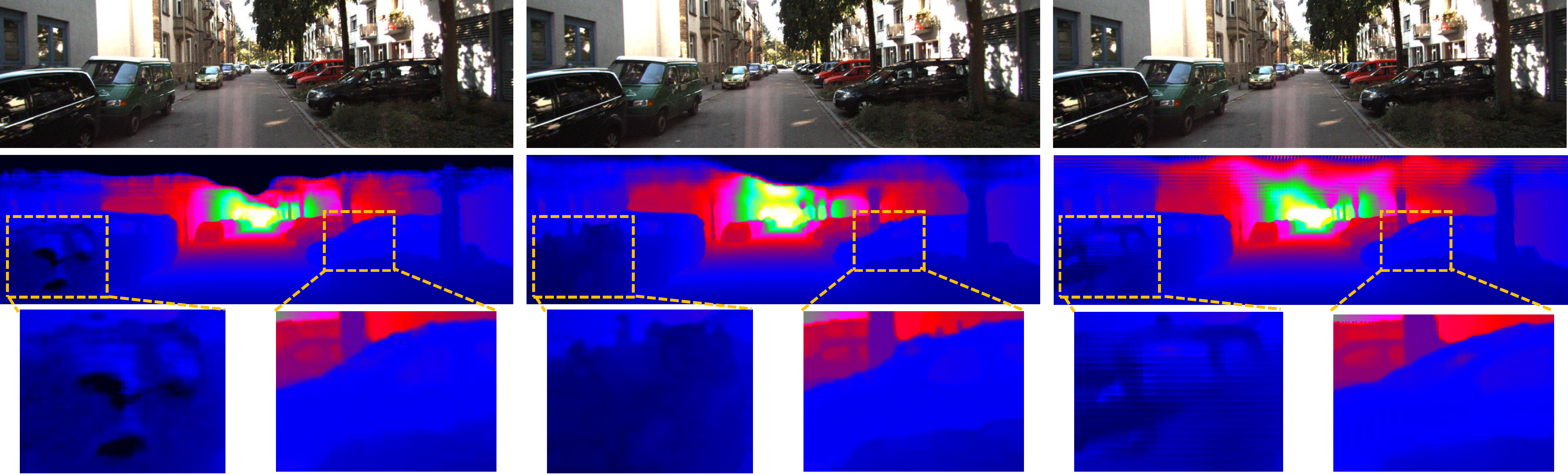}}
		\vspace{-0.2em}
		\\
		\multicolumn{3}{c}{\includegraphics[width=0.99\textwidth]{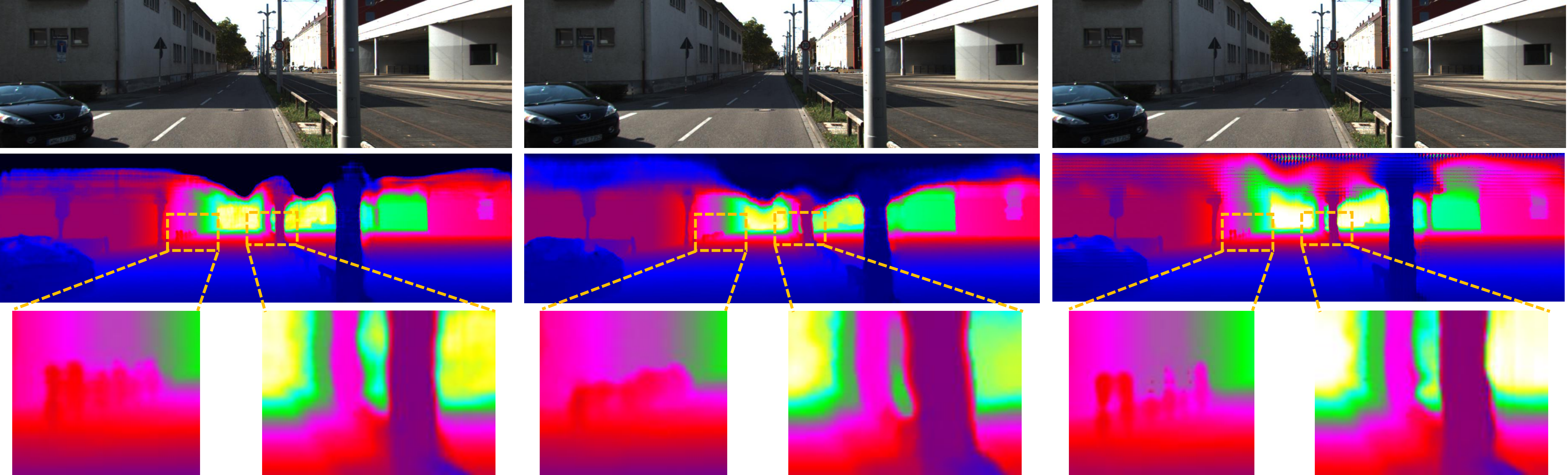}}
		\\
		\hspace{30pt}  \small{(a) Sparse-to-Dense \cite{ma2018self}} \hspace{90pt}  & \small{(b) CSPN \cite{cheng2018depth}} \hspace{45pt} & \small{(c) Our method} \\
	\end{tabular}
	\caption{{\bf Qualitative comparison on KITTI test set.} We show results of our method and top ranking methods: CSPN \cite{cheng2018depth} and Sparse-to-Dense \cite{ma2018self}. For each example, we show color image, dense depth output, and zoom-in view of some local areas. Our model produces more accurate results aligning better with the color image. Our model also preserves thin structures like tree, traffic light, and road lamp.}
	\vspace{-1.5em}
	\label{fig:kitti_test_quality}
\end{figure*}

\vspace{-0.2em}
\noindent\textbf{Evaluate on KITTI Test Set.}
We first evaluate our method on the test set of the KITTI depth completion benchmark.
The test set contains 1000 data, including color image, sparse LiDAR depth, and transformation between color camera and LiDAR. 
Ground truth are held, and evaluation can be only done on their server to prevent overfitting.
The evaluation server calculates four metrics: root mean squared error (RMSE mm), mean absolute error (MAE mm), root mean squared error of the inverse depth (iRMSE 1/km) and mean absolute error of the inverse depth (iMAE 1/km), among which RMSE is the most important indicator and chosen to rank submissions on the leader-board since it measures error directly on depth and penalizes on further distance where depth measurement is more challenging.



The performances of our methods and all the other high ranking methods are listed in  \tabref{tab:kitti_test}.
Our method ranks the 1st on the leader-board at the time of submission, outperforming the 2nd with significant improvement.
Qualitative comparison with some competing methods \cite{ma2018self, cheng2018depth} are shown in \figref{fig:kitti_test_quality}.
For each example, we show both the recovered complete depth, together with zoom-in view to high-light some details.
In general, our method produces more accurate depth (e.g., the complete car) with better details (e.g., road-side railing).
The running time of our model on a single GPU (Nvidia GTX 1080Ti) is 0.07s per image.

\begin{figure*}[t]
	\centering
	\begin{tabular}{*{4}{c@{\hspace{1.8px}}}}
		\raisebox{0.5\height}{\rotatebox{90}{\small{~RGB}}} & \includegraphics[width=0.31\textwidth]{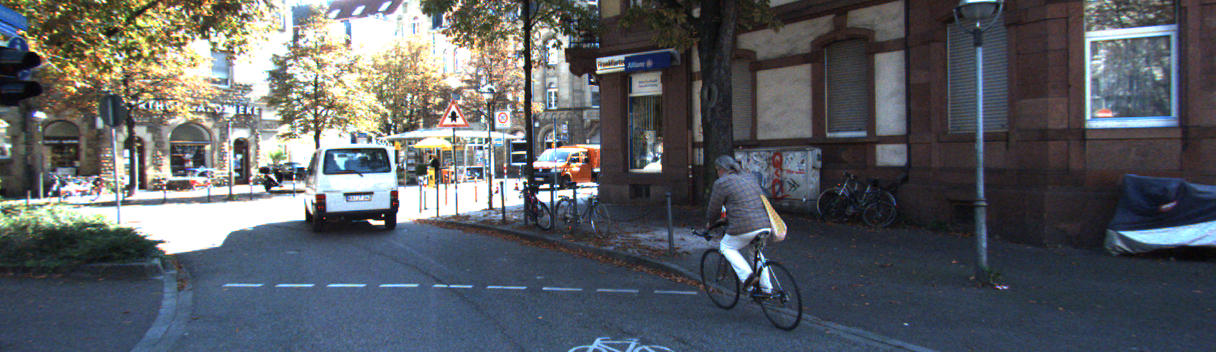}&
		\includegraphics[width=0.31\textwidth]{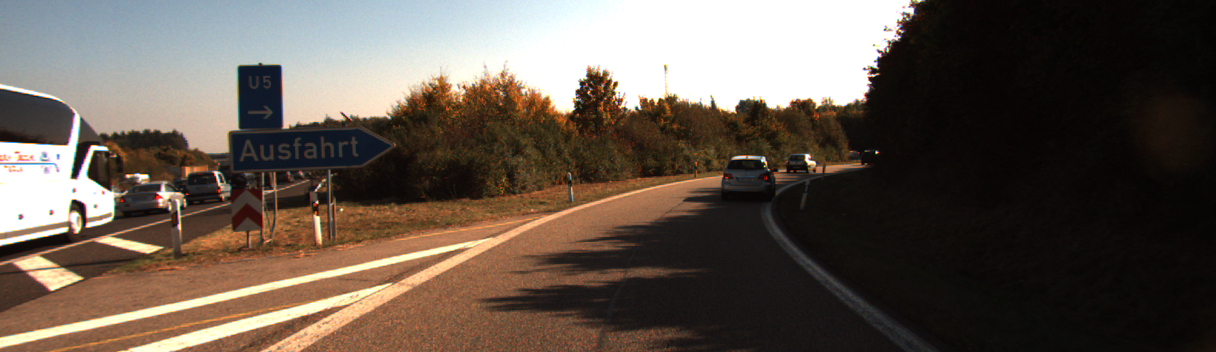} &
		\includegraphics[width=0.31\textwidth]{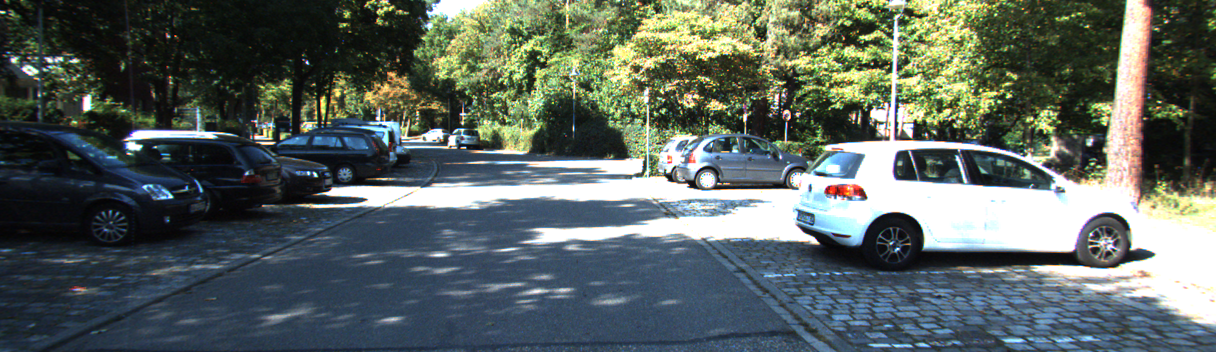}
		\vspace{-1.5pt}
		\\
		\vspace{-1.5pt}
		\raisebox{0.5\height}{\rotatebox{90}{\small{Sparse}}} & \includegraphics[width=0.31\textwidth]{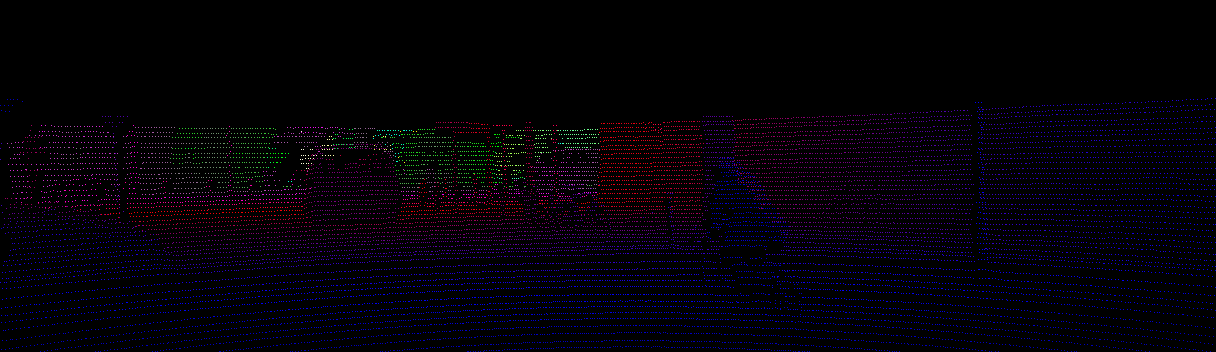}&
		\includegraphics[width=0.31\textwidth]{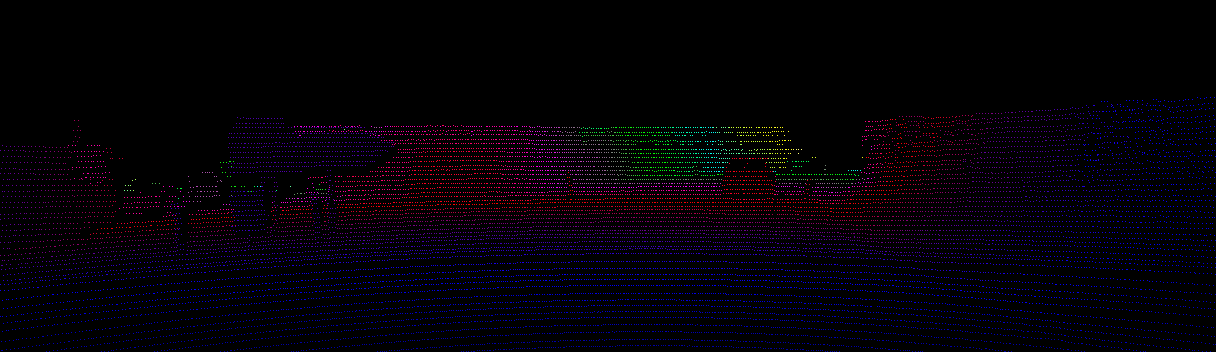}&
		\includegraphics[width=0.31\textwidth]{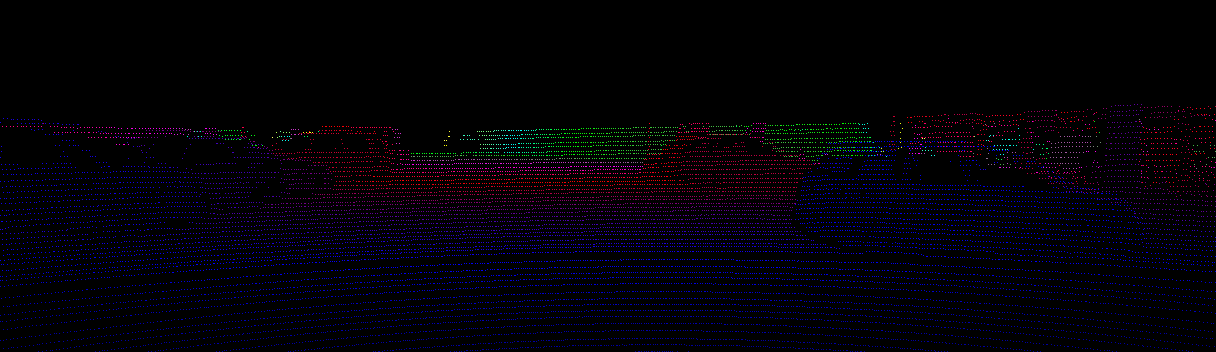}\\
		\vspace{-1.5pt}
		\raisebox{0.0\height}{\rotatebox{90}{\small{~Confidence}}} &\includegraphics[width=0.31\textwidth]{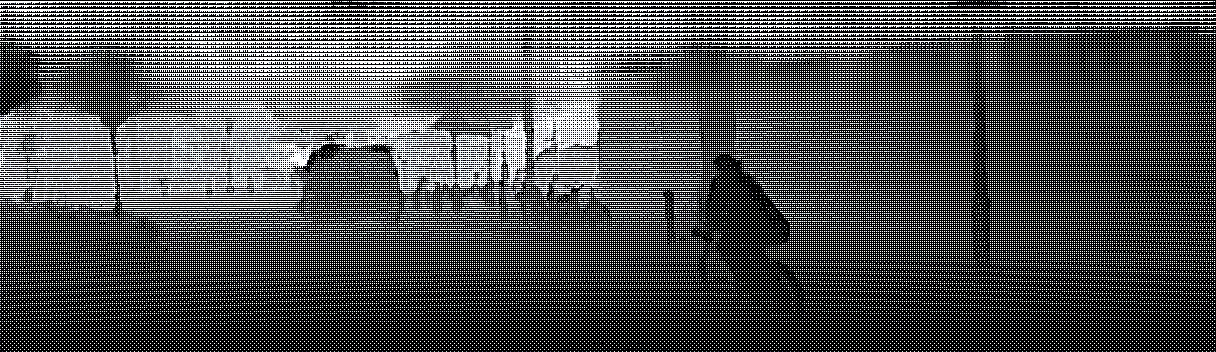}&
		\includegraphics[width=0.31\textwidth]{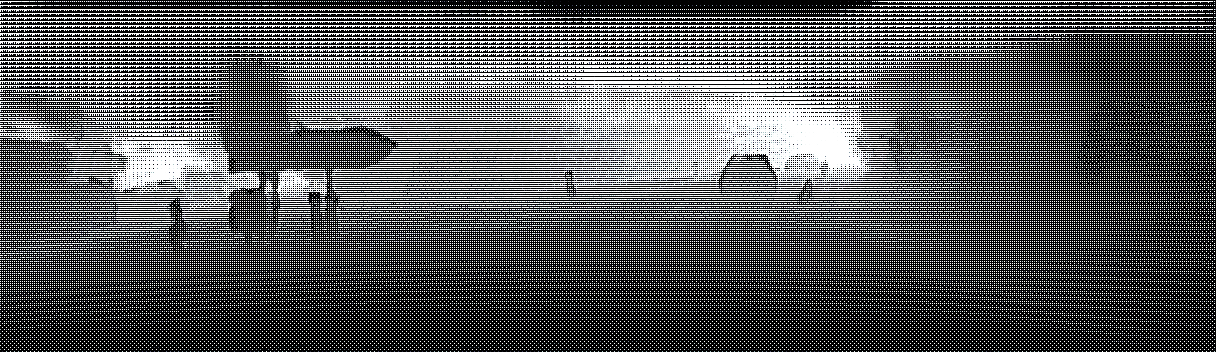} &
		\includegraphics[width=0.31\textwidth]{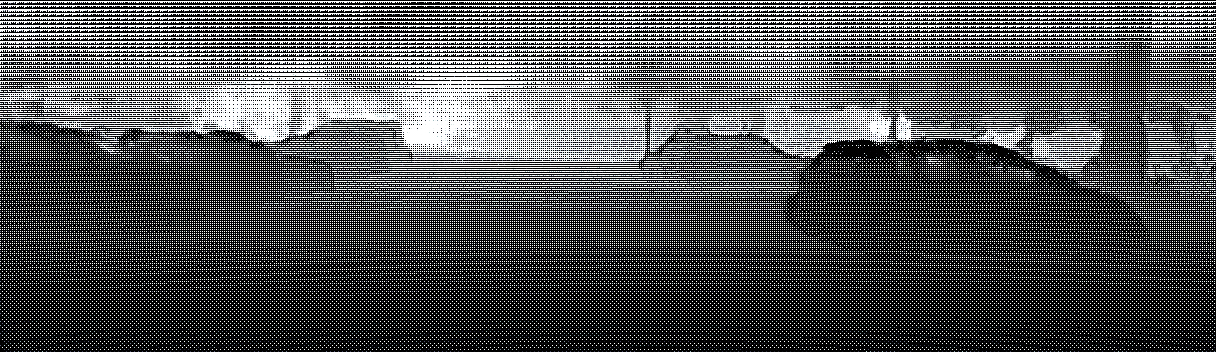}\\
		\vspace{-1.5pt}
		\raisebox{0.\height}{\rotatebox{90}{\small{~~~~Normal}}} & \includegraphics[width=0.31\textwidth]{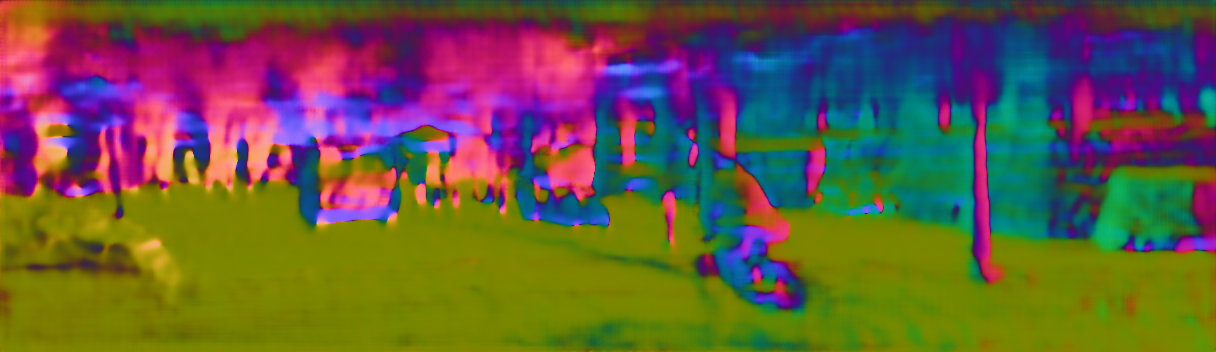}&
		\includegraphics[width=0.31\textwidth]{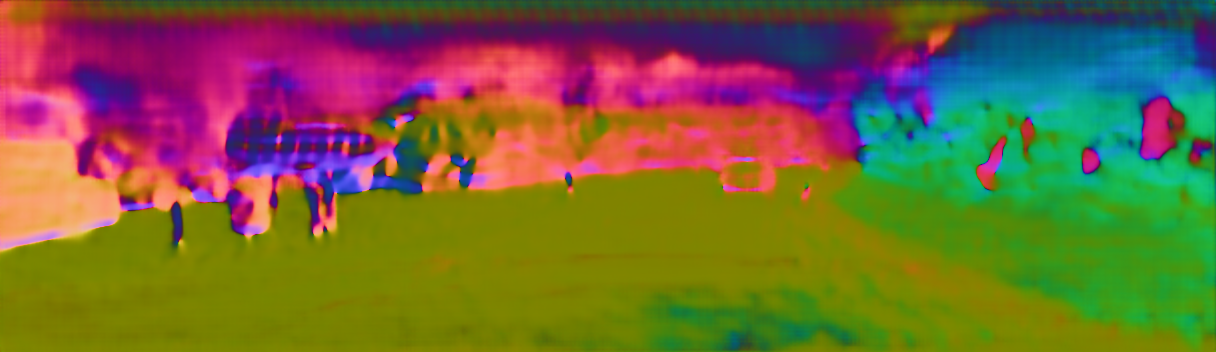} &
		\includegraphics[width=0.31\textwidth]{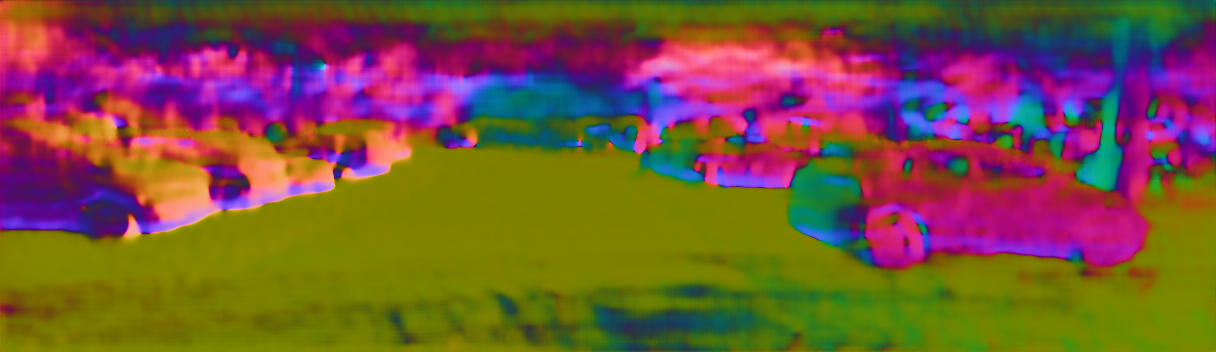}\\
		\vspace{-1.5pt}
		\raisebox{1.5\height}{\rotatebox{90}{$w_c$}} & \includegraphics[width=0.31\textwidth]{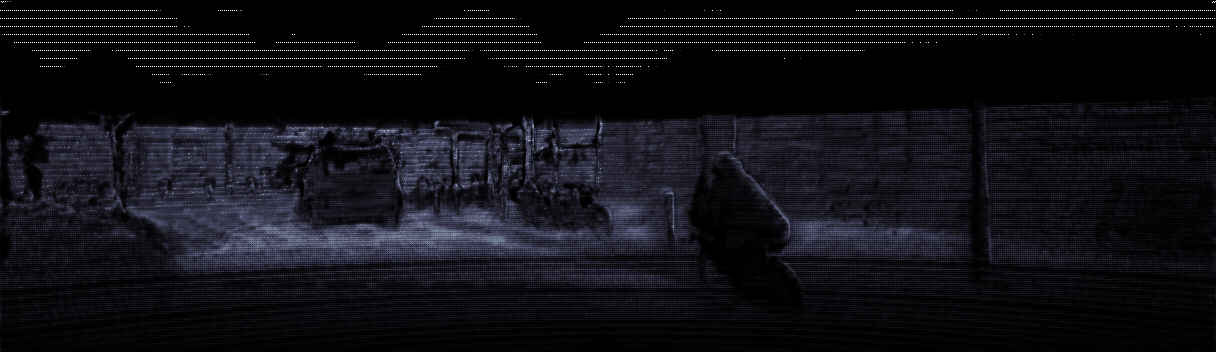}&
		\includegraphics[width=0.31\textwidth]{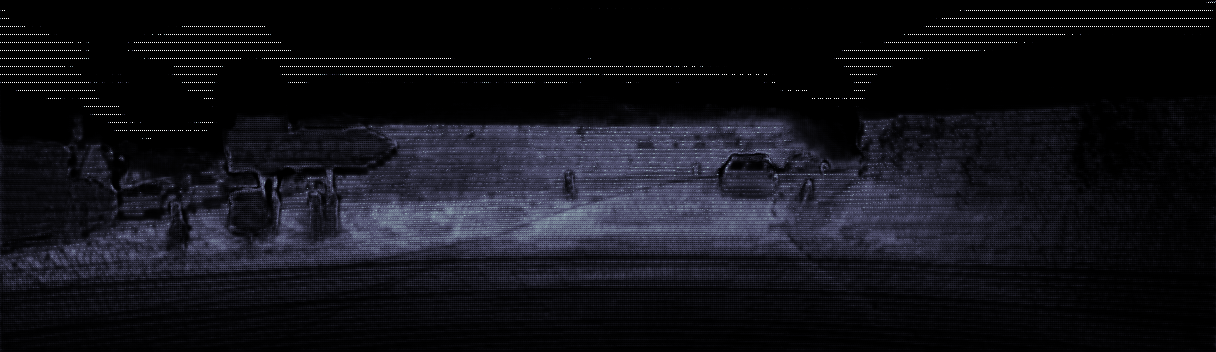} &
		\includegraphics[width=0.31\textwidth]{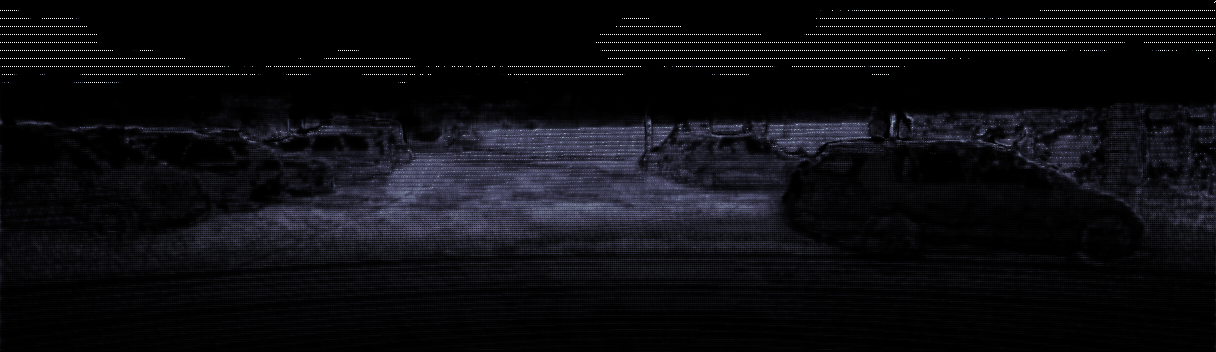}\\
		\vspace{-1.5pt}
		\raisebox{1.0\height}{\rotatebox{90}{$w_n$}} & \includegraphics[width=0.31\textwidth]{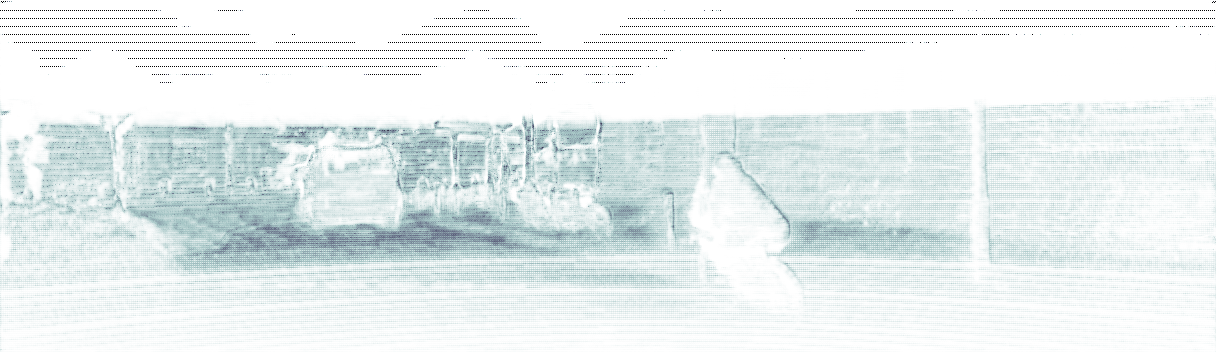}&
		\includegraphics[width=0.31\textwidth]{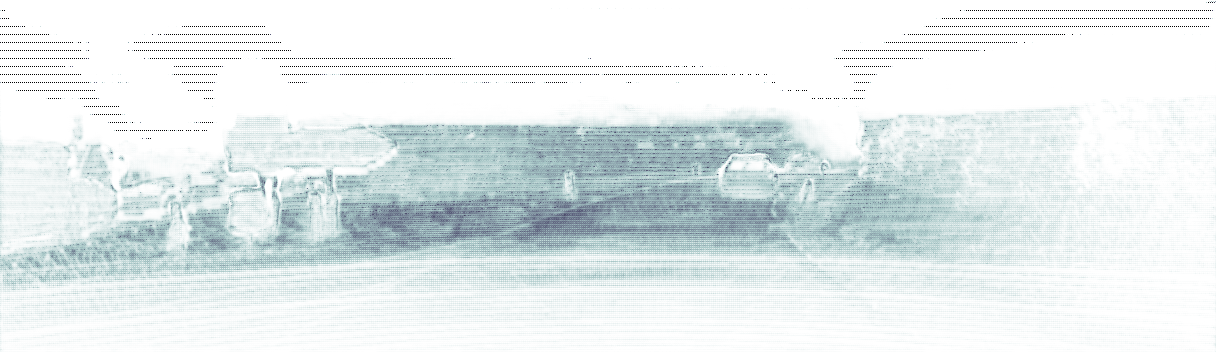} &
		\includegraphics[width=0.31\textwidth]{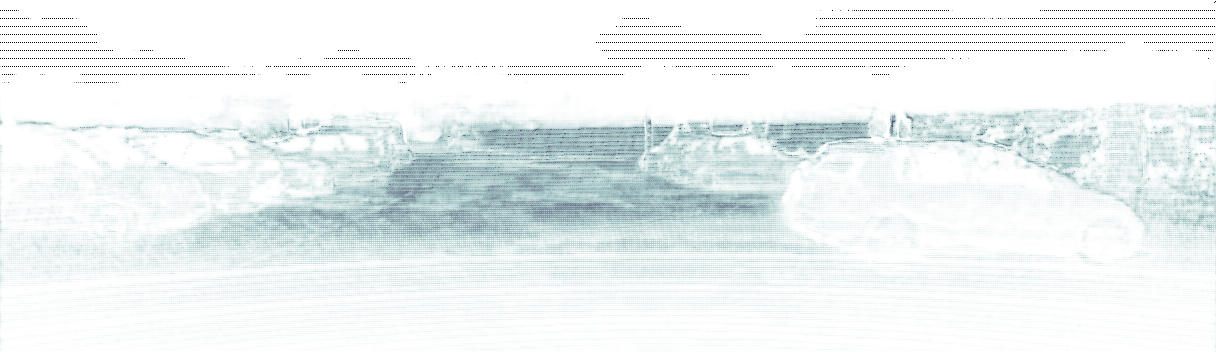}\\
		\vspace{-1.5pt}
		\raisebox{0.0\height}{\rotatebox{90}{\small{~~~Bilateral }}} &  \includegraphics[width=0.31\textwidth]{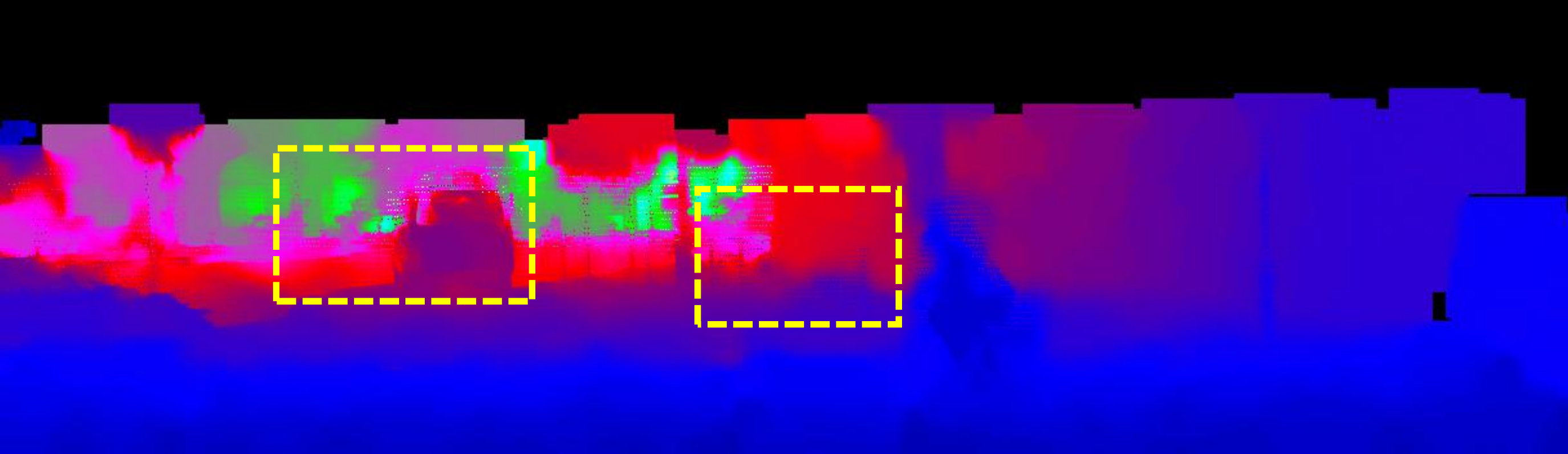}&
		\includegraphics[width=0.31\textwidth]{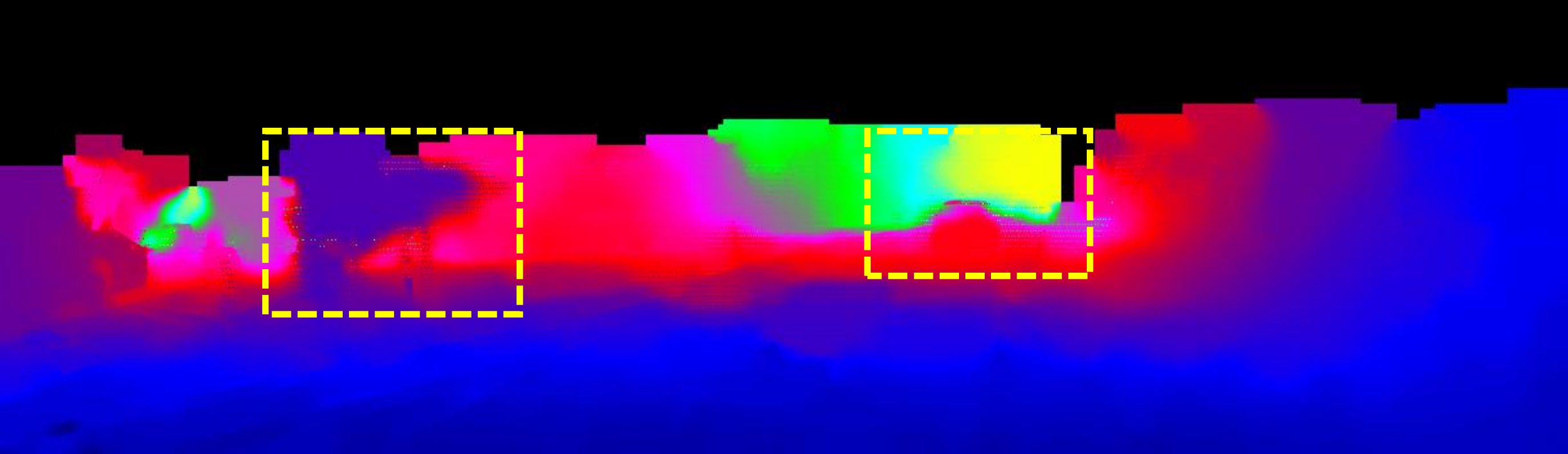}&
		\includegraphics[width=0.31\textwidth]{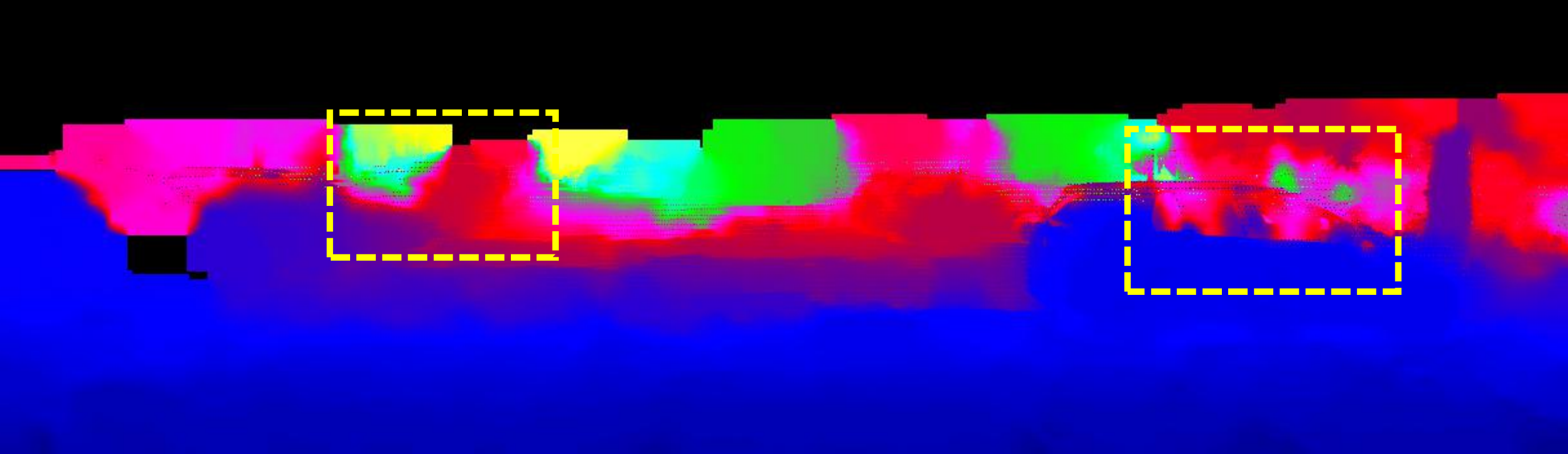}\\
		\vspace{-1.5pt}
		\raisebox{0.0\height}{\rotatebox{90}{\small{~~~~~~~~Fast }}} & \includegraphics[width=0.31\textwidth]{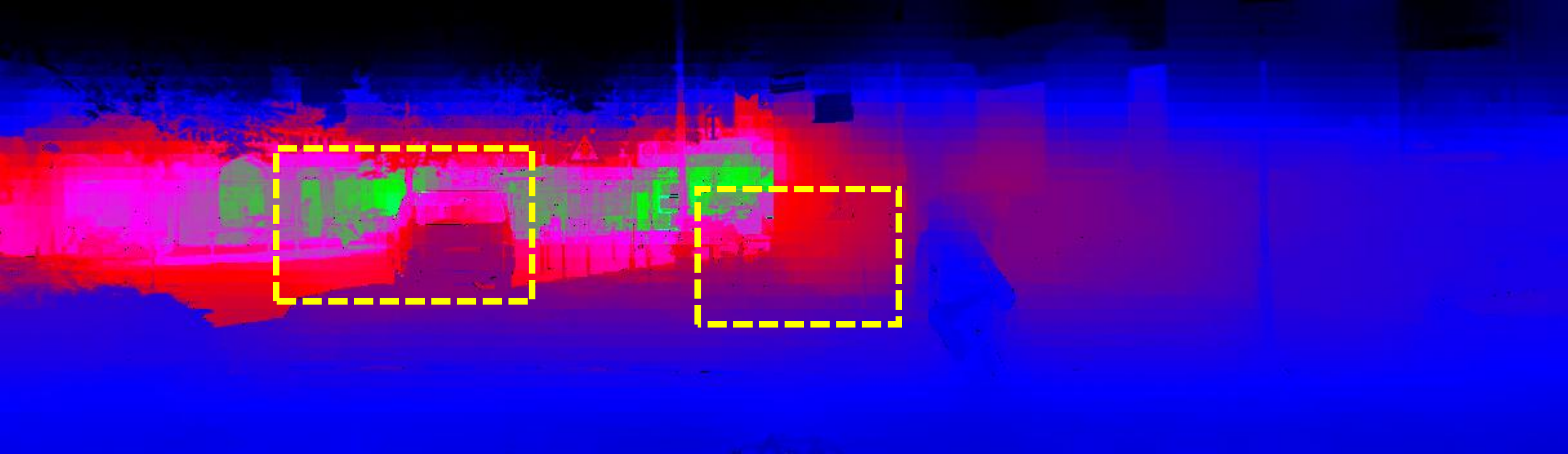}&
		\includegraphics[width=0.31\textwidth]{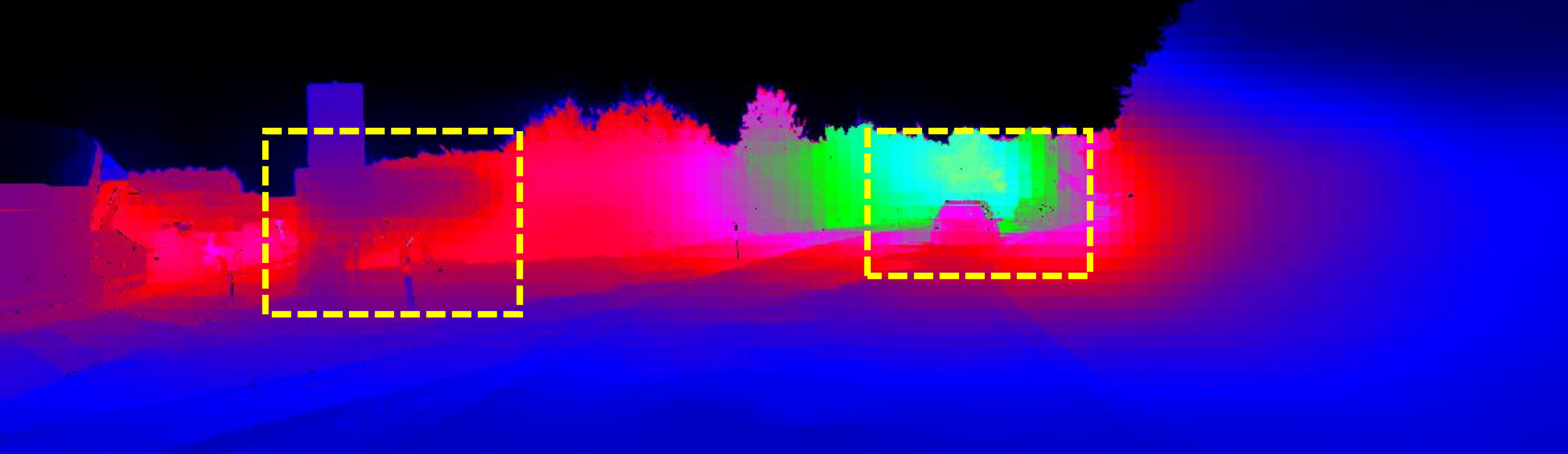}&
		\includegraphics[width=0.31\textwidth]{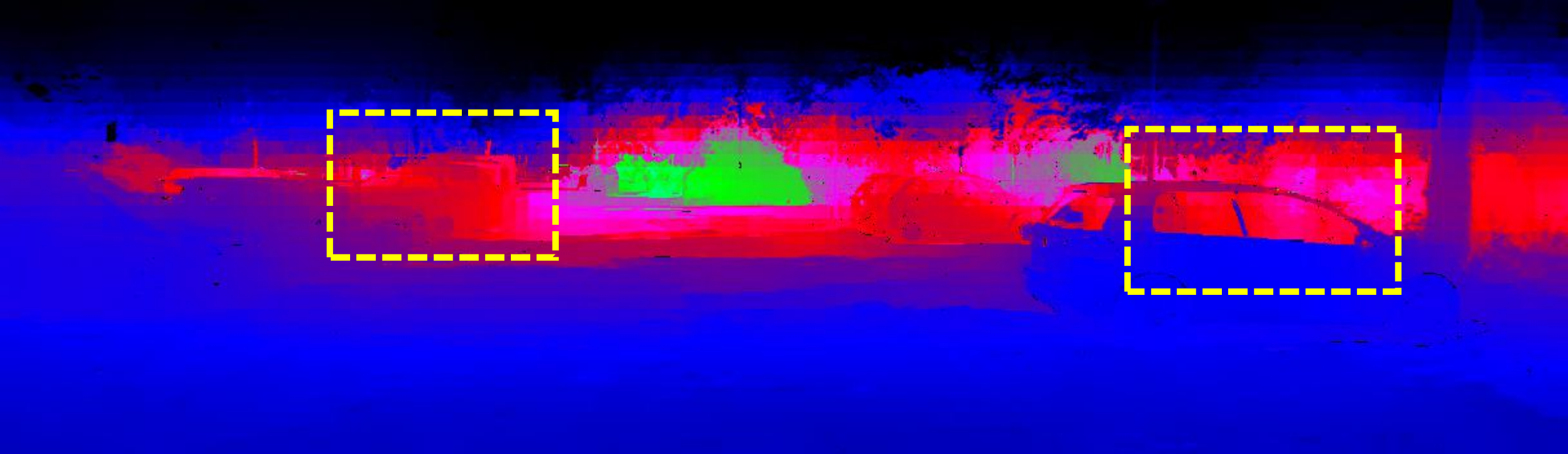}\\
		\vspace{-1.5pt}
		\raisebox{0.5\height}{\rotatebox{90}{\small{~~TGV }}} & \includegraphics[width=0.31\textwidth]{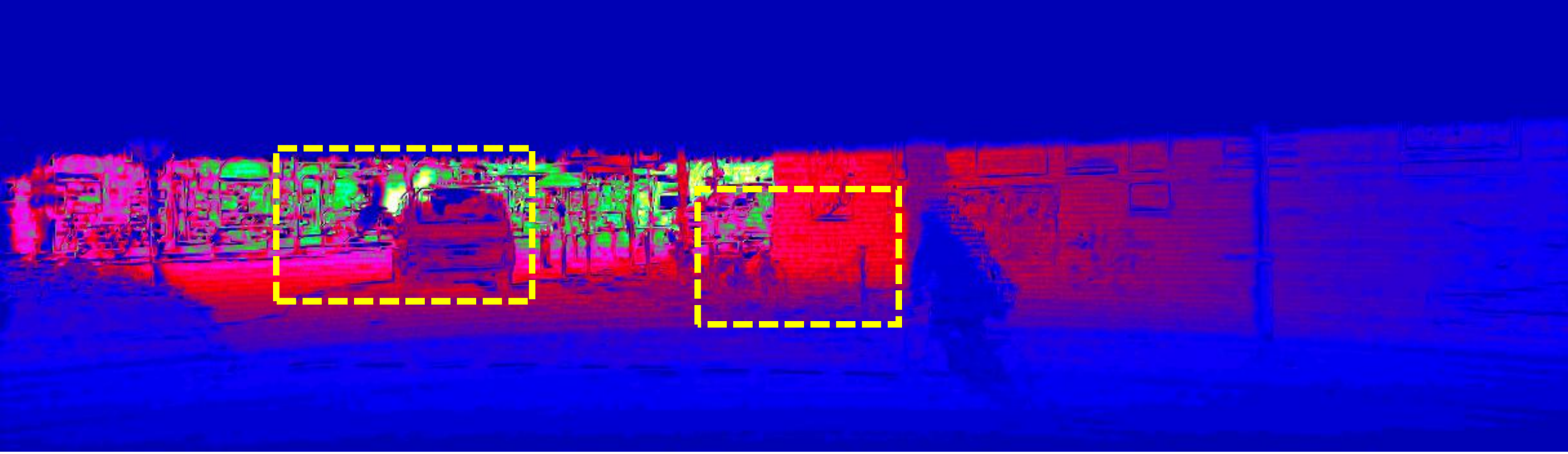}&
		\includegraphics[width=0.31\textwidth]{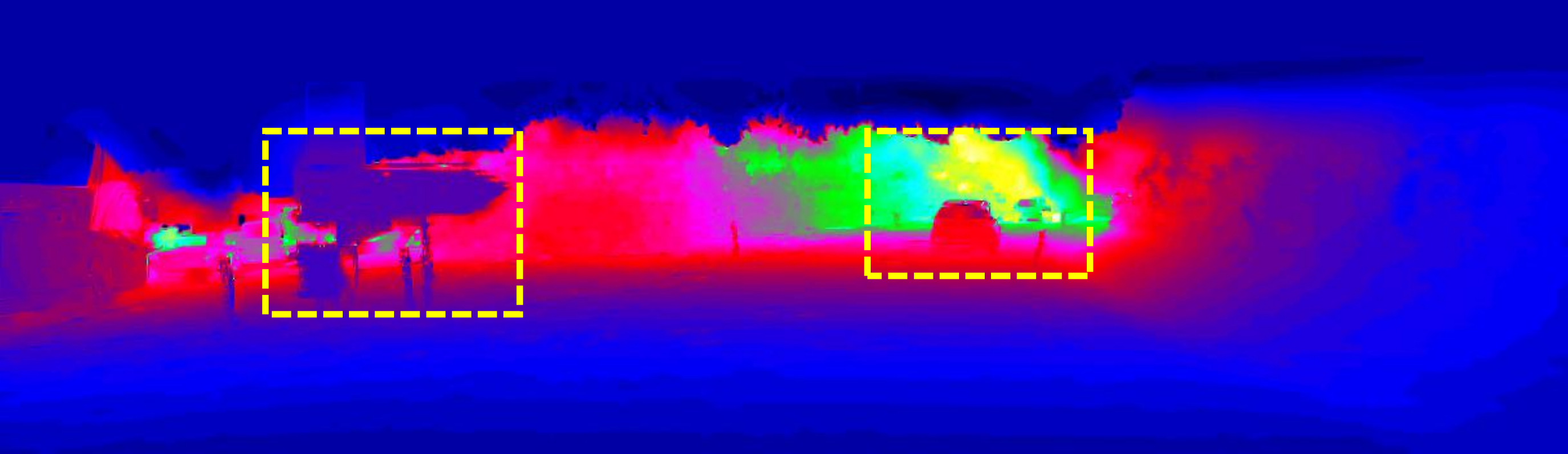}&
		\includegraphics[width=0.31\textwidth]{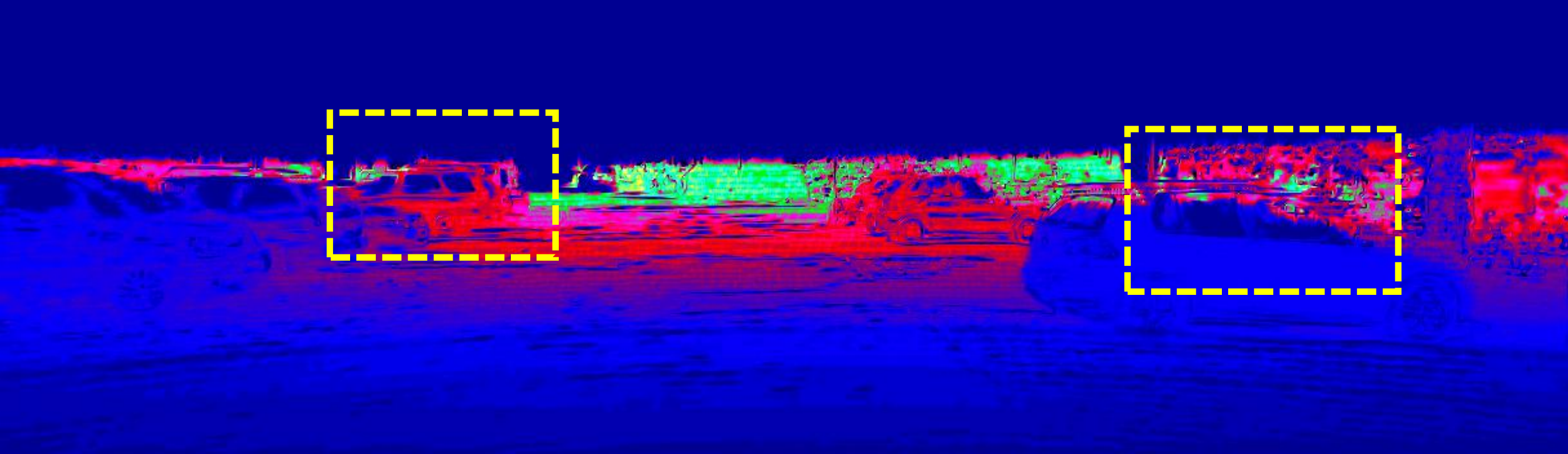}\\
		\vspace{-1.5pt}
		\raisebox{0.0\height}{\rotatebox{90}{\small{Zhang \etal}}} & \includegraphics[width=0.31\textwidth]{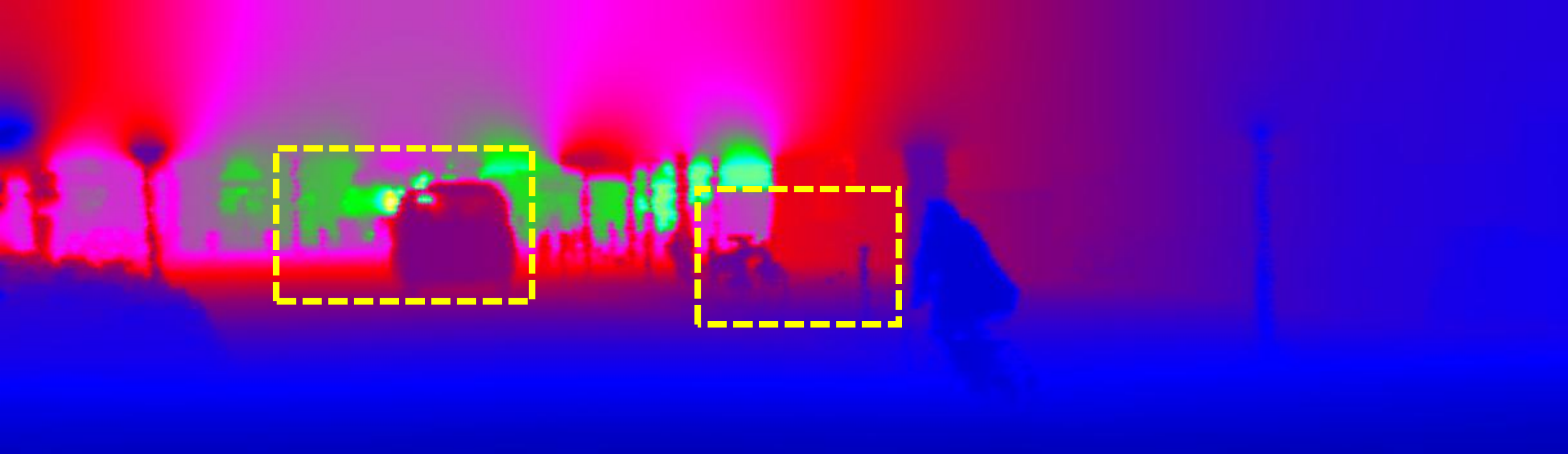}&
		\includegraphics[width=0.31\textwidth]{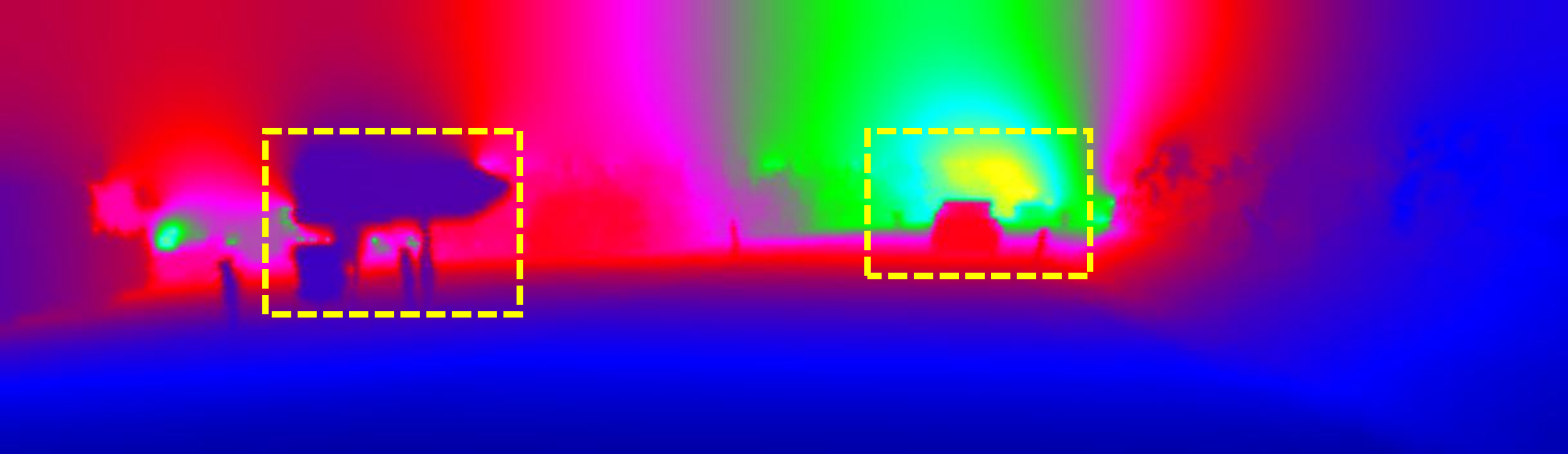}&
		\includegraphics[width=0.31\textwidth]{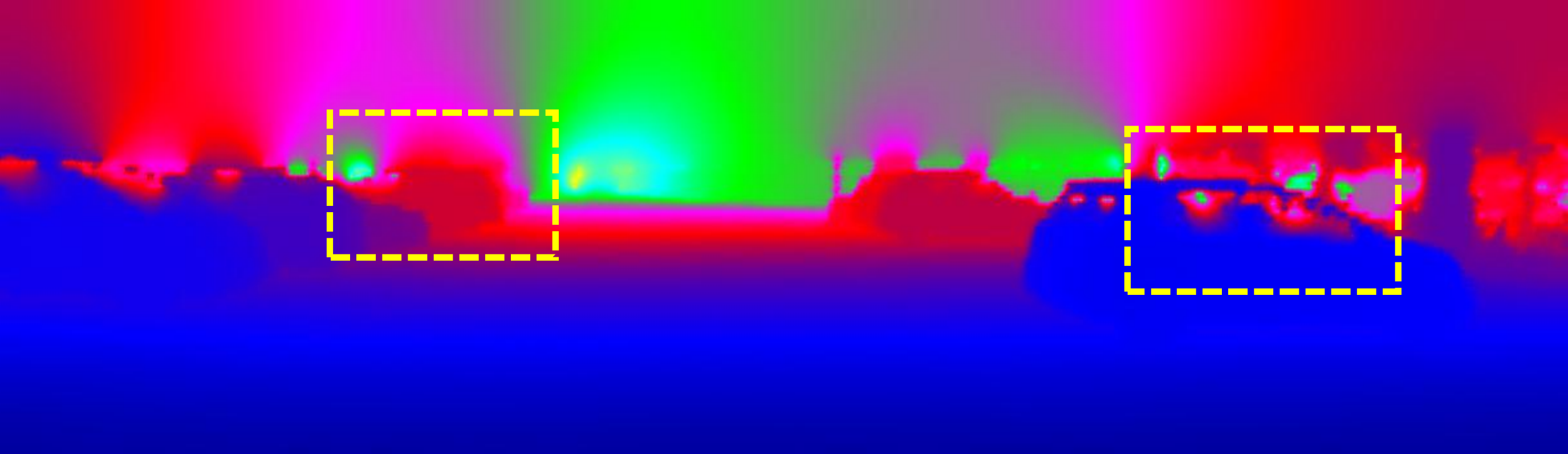}\\
		\vspace{-1.5pt}
		\raisebox{0.0\height}{\rotatebox{90}{\small{~~~~~~~Ours}}} & \includegraphics[width=0.31\textwidth]{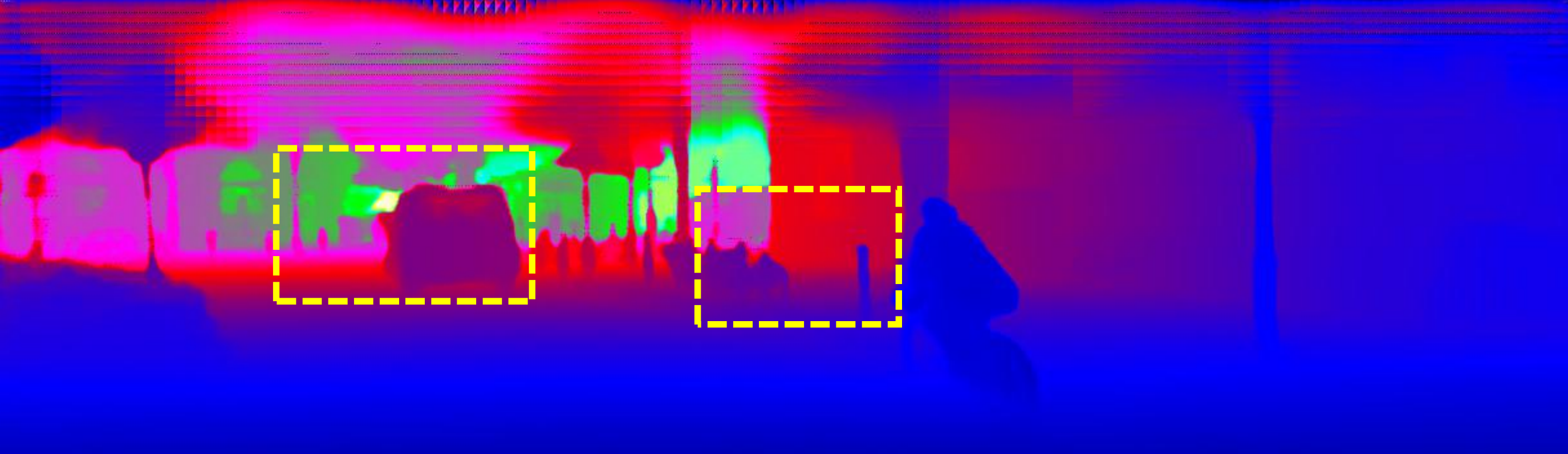}&
		\includegraphics[width=0.31\textwidth]{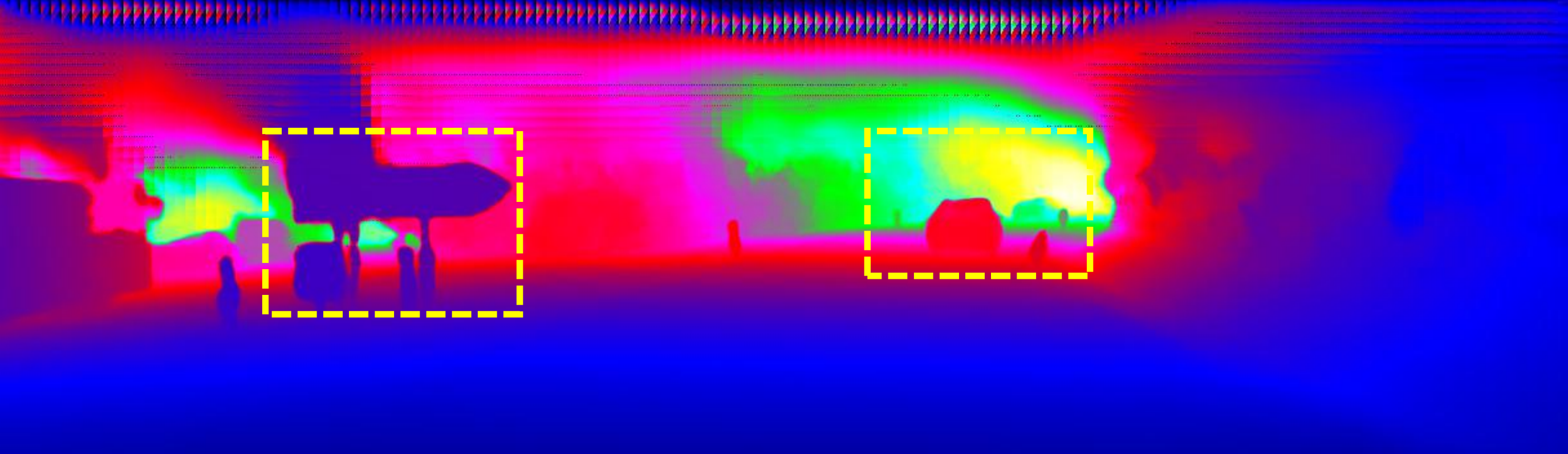}&
		\includegraphics[width=0.31\textwidth]{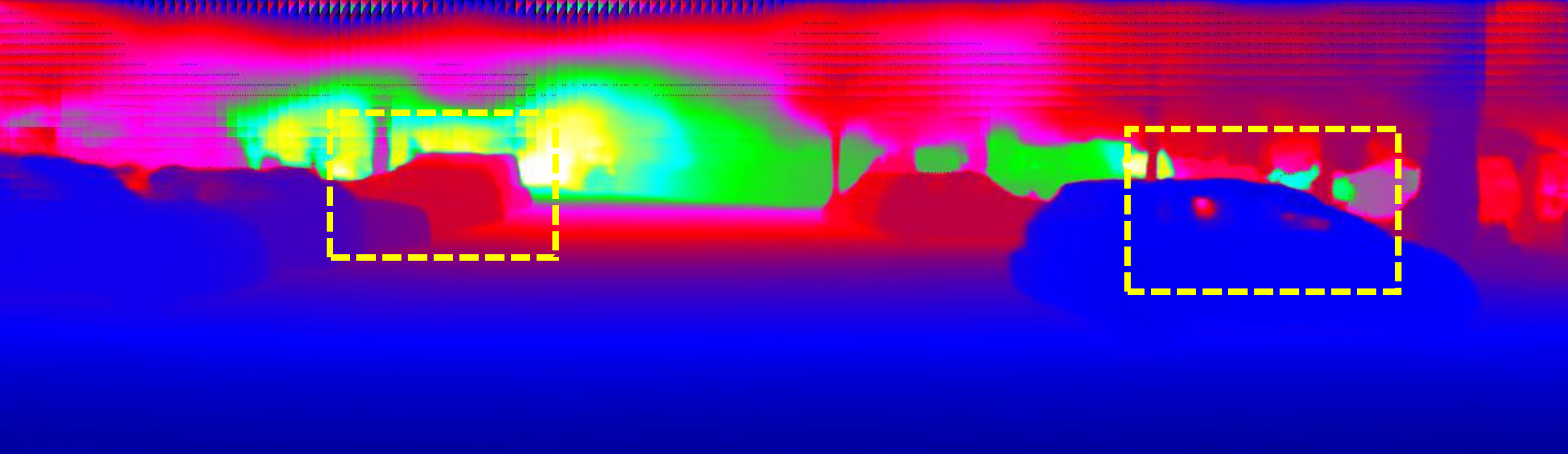}\\
	\end{tabular}
	\caption{{\bf Qualitative results on KITTI validation set.} From top to bottom are RGB image input, sparse depth input, confidence mask, estimated surface normals, attention map for color pathway, attention map for normal pathway, results of Bilateral \cite{silberman2012indoor}, Fast \cite{barron2016fast}, TGV \cite{ferstl2013image} Zhang \etal~\cite{zhang2018deepdepth}, and our method. We mark some regions in the results to highlight the difference across methods. }
	\label{fig:kitti_valid_quality}
	\vspace{-0.8em}
\end{figure*}

{\setlength{\tabcolsep}{0.5em}
	\begin{table}[t]
		\centering \footnotesize
		\begin{tabular}{l|c|c|c|c}
			\hline
			& \textbf{RMSE} & MAE & iRMSE & iMAE \\ \cline{1-4}
			\hline
			CSPN~\cite{cheng2018depth} & 1019.64 & 279.46 & 2.93 & 1.15 \\
			Spade-RGBsD~\cite{jaritz2018sparse} & 917.64 & 234.81 & \textbf{2.17} & \textbf{0.95} \\
			HMS-Net~\cite{huang2018hms} & 841.78 & 253.47 & 2.73 & 1.13  \\
			MSFF-Net~\cite{wang2018multi} & 836.69 & 241.54 & 2.63 & 1.07 \\
			NConv-CNN~\cite{eldesokey2018propagating} & 829.98 & 233.26 & 2.60 & 1.03 \\
			Sparse-to-Dense~\cite{ma2018self} & 814.73 & 249.95 & 2.80 & 1.21 \\
			\hline
			Ours & \textbf{758.38} & \textbf{226.50} & 2.56 & 1.15 \\
			\hline
		\end{tabular}
		\vspace{0.5em}
		\caption{{\bf Performance of depth completion on KITTI test set \cite{uhrig2017sparsity}.} The evaluation is done via KITTI testing server, and submissions are ranked by RMSE on the leaderboard. Our model outperforms the 2nd \cite{ma2018self} on RMSE with a large gap.
		}
		\vspace{-1em}
		\label{tab:kitti_test}
	\end{table}
}

{\setlength{\tabcolsep}{0.5em}
	\begin{table}[t]
		\centering \footnotesize
		\begin{tabular}{l|c|c|c|c}
			\hline
			& \textbf{RMSE} & MAE & iRMSE & iMAE \\ \cline{1-4}
			\hline
			Bilateral \cite{silberman2012indoor} & 2989.02 & 1200.56 & 9.67 & 5.08 \\
			Fast \cite{barron2016fast} & 3548.87 & 1767.80 & 26.48 & 9.13 \\
			TGV \cite{ferstl2013image} & 2761.29 & 1068.69 & 15.02 & 6.28 \\
			Zhang \etal~\cite{zhang2018deepdepth} &1312.10  &356.60  & 4.29  & 1.41 \\
			\hline
			Ours &  \textbf{687.00} & \textbf{215.38} & \textbf{2.51} & \textbf{1.10} \\
			\hline
		\end{tabular}
		\vspace{0.5em}
		\caption{{\bf Performance of depth completion on KITTI validation set \cite{uhrig2017sparsity}.} We compare to non-learning based approaches \cite{silberman2012indoor,barron2016fast,ferstl2013image} and Zhang \etal \cite{zhang2018deepdepth}. Our method performs the best on all the evaluation metrics.}
		\label{tab:kitti_valid}
	\end{table}
}

\noindent\textbf{Evaluation on KITTI Validation Set.}
We further compare on the validation set of KITTI benchmark to other related methods that are not on the benchmark, including bilateral filter using color (Bilateral), fast bilateral (Fast), optimization using total variance (TGV), and deep depth completion for indoor scene \cite{zhang2018deepdepth}.
Models are trained on the training set only.
The quantitative results are shown in \tabref{tab:kitti_valid}.
As can be seen, our method significantly outperforms all the other methods.
Non-learning based approaches \cite{silberman2012indoor, barron2016fast, ferstl2013image} do not perform well possibly because of drastic illumination change and complicate scene structures.
Zhang \etal \cite{zhang2018deepdepth} performs much better than the above mentioned methods but still far from our model as it does not handle outdoor specific issues.

Qualitative comparisons are shown in \figref{fig:kitti_valid_quality}. From the highlighted regions, the Bilateral \cite{silberman2012indoor} and Fast \cite{barron2016fast} over-smooth the boundaries and details of objects. In contrast, TGV \cite{ferstl2013image} generates the detailed structures, but noisy smooth surfaces, like roads. 
Zhang \etal \cite{zhang2018deepdepth} performs well on close regions, but worse than our method in far areas and where surface normal estimation fails, e.g., traffic sign and car windows. 
Our method successfully solves these problems for two reasons. Firstly, we integrate the offline linear optimization into network, which allows end-to-end training for presumably more optimal solutions. From \tabref{tab:ablation} (``-Attention Integration''), we can see that the depth prediction from our normal pathway is already much better than Zhang \etal \cite{zhang2018deepdepth}. 
Secondly, we further learn a confidence mask to handle occlusion and use the attention based integration to improve the area where normal pathway fails.

\vspace{-0.2em}
\subsection{Ablation Study}
\vspace{-0.2em}
To understand the impact of each model components on the final performance, we conduct a comprehensive ablation study by disabling each component respectively and show how result changes.
Quantitative results are shown in \tabref{tab:ablation}. Performance drops reasonably with each component disabled, and the full model works the best.

\vspace{2.0pt}
\noindent\textbf{Effect of Surface Normal Pathway.}
To verify if surface normal is a reasonable intermediate depth representation for outdoor scene similar as for the indoor case, we train a model without estimating the normal but directly output the complete depth.
Under this setting, there is also no attention integration since only one pathway is available.
The performance is shown as ``-Normal Pathway'' in \tabref{tab:ablation}. The performance drops significantly, i.e. RMSE increases 
about 87mm, compared to our full model.
This demonstrates that surface normal is also helpful for outdoor depth completion.

{\setlength{\tabcolsep}{0.5em}
	\begin{table}[t]
		\centering \footnotesize
		\begin{tabular}{l|c|c|c|c}
			\hline
			~~Models & \textbf{RMSE} & MAE & iRMSE & iMAE \\ \cline{1-4}
			\hline
			- Normal Pathway & 774.25 & 258.77 & 4.65 & 1.40 \\
			- Attention Integration & 729.96 & 239.08 & 2.74 & 1.20 \\
			- DCU & 767.82 & 246.36 & 2.69 & 1.17 \\
			- Confidence mask & 756.32 & 272.91 & 2.70 & 1.19 \\
			\hline
			~~Full & \textbf{687.00} & \textbf{215.38} & \textbf{2.51} & \textbf{1.10} \\
			\hline
		\end{tabular}
		\vspace{0.5em}
		\caption{{\bf Ablation study of depth completion on KITTI validation set.} We re-train our model with each major component disabled and evaluate on KITTI validation set. Our full model with all the components on achieves the best performance. }
		\label{tab:ablation}
		\vspace{-0.5em}
	\end{table}
}

\vspace{2.0pt}
\noindent\textbf{Effect of Attention Based Integration.}
We then disable the attention based integration to verify the necessity of the two-pathway combination,
\ie only considering the depth from the normal pathway.
Without this integration, all the evaluation metrics drop (\tabref{tab:ablation} ``-Attention Integration'') compared to the full model.
\figref{fig:kitti_valid_quality} (row $w_c$, $w_n$) shows the attention map learned automatically for color pathway and surface normal pathway. It can be seen that surface normal pathway works better (i.e. higher weight) in close range but gets worse when the distance goes up, which is consistent with our analysis.
In contrast, the color pathway cannot capture accurate details in close range compared to the surface normal pathway but better in far distance.
Although the color pathway works better for fewer regions compared to the surface normal pathway, it is critically important to achieve good performance in far area, where large error are more likely to happen.

\noindent\textbf{Effect of Deep Completion Unit.}
We also replace our deep completion unit to a traditional encoder-decoder architecture with early fusion, where the input color image, sparse depth, and a binary mask are concatenated at the beginning and fed as input to the network.
This modification causes significant performance drop even with all the other components of the model enabled (\tabref{tab:ablation} ``-DCU'').
Notice that we sum the features from the sparse depth encoder with the features from decoder rather than the ordinary concatenation. 
We also tried the concatenation option which however takes more memory and produces slightly worse performance. 

\begin{figure}[t]
\vspace{-4mm}
	\centering
	\begin{tabular}{*{2}{c@{\hspace{0px}}}}
		\hspace{-1.5em} \includegraphics[width=0.25\textwidth]{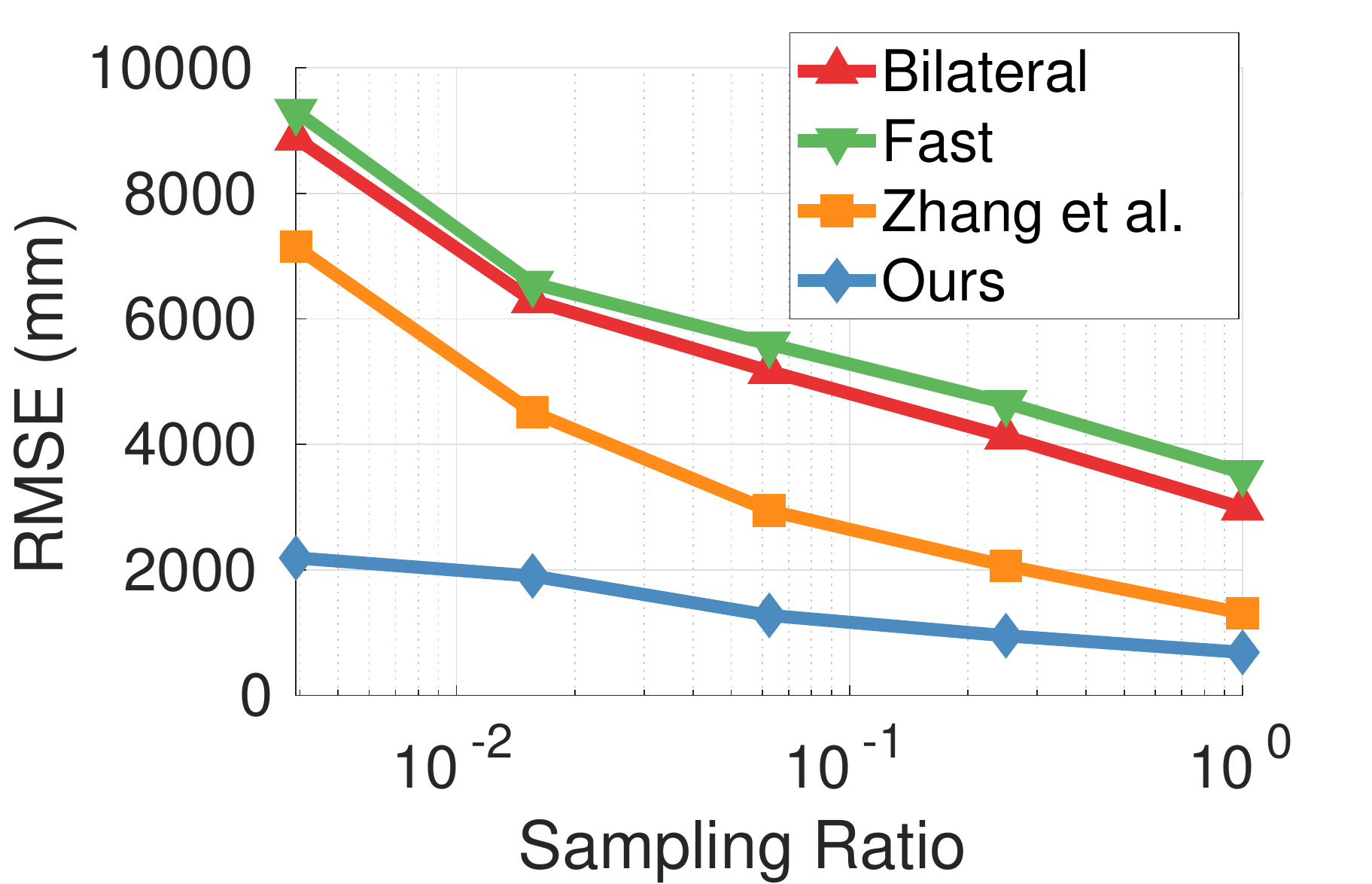} &
		\includegraphics[width=0.25\textwidth]{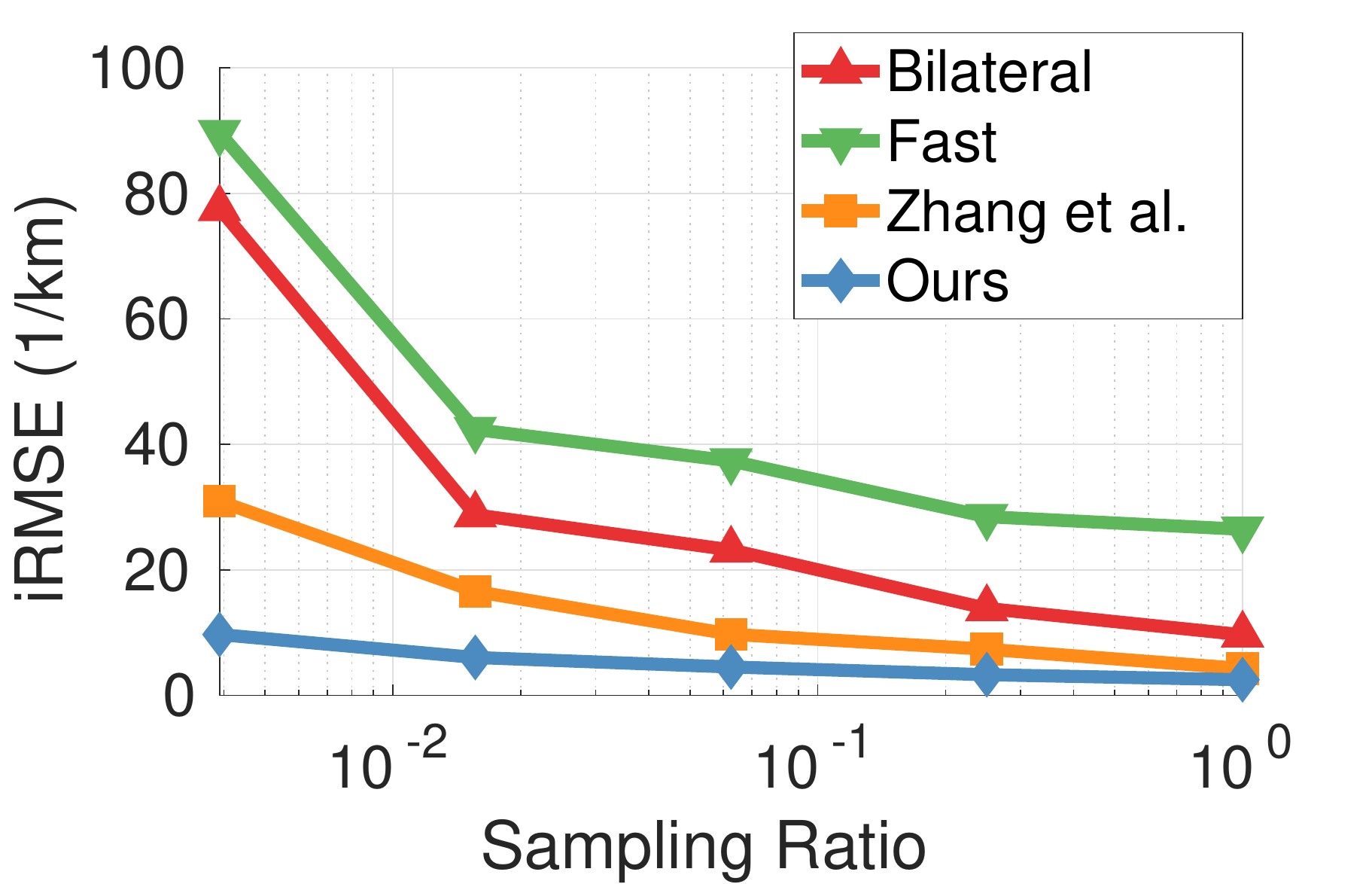} \\
	\end{tabular}
	\vspace{-0.5em}
	\caption{{\bf Performance with different sparsity.} We test our model on input depth with different sparsity by sub-sampling the raw LiDAR.
	Our method works well even with very sparser depth input, and outperforms the other methods.
		}
	\label{fig:sparsity}
	\vspace{-0.2em}
\end{figure}

\vspace{2.0pt}
\noindent\textbf{Effect of Confidence Mask.}
Last but not least, we disable the confidence mask by replacing the learned one with a typical binary mask indicating the availability of sparse depth per-pixel.
This causes dramatic increase of RMSE by 69mm compared to the full model.
In contrast, our full model learns confidence masks according to the inputs, which provide extremely useful information about the reliability of the input sparse depth for the surface normal pathway, as shown in \figref{fig:misalignemnt}(d) and \figref{fig:kitti_valid_quality}(Confidence).
As can be seen, the area with overlapping depth from foreground and background are generally marked with low confidence.
Notice that these areas usually happens on the boundary of the foreground where occlusion happens.

\vspace{-0.2em}
\subsection{Generalization Capability}
\vspace{-0.2em}
Even though in this paper we especially focus on producing dense depth for car-held LiDAR devices, our model can be considered as a general depth completion approach.
Therefore, we investigate the generalization capability of our model under different situations, specifically with different input depth sparsity and in indoor environment.

\vspace{2.0pt}
\noindent\textbf{Robustness Against Depth Sparsity.}
It is interesting to see if our model could still work on more challenging cases where input depths are even sparser.
The raw LiDAR depth provided by the benchmark is roughly 18,400 samples per depth image of 1216 by 352 resolution, i.e. 4.3\% of the pixels having depth.
We uniformly sub-sample the raw LiDAR depth by ratios of $1/4$, $1/16$, $1/64$, and $1/256$, which correspond to 1.075\%, 0.269\%, 0.0672\%, and 0.0168\% of pixels having depth.
It worth noting that 0.0168\% corresponds to 72 pixels per depth image.
This is an extreme hard case where the scene structure is almost missing from the input sparse depth.


The performances of our model and other methods  \cite{silberman2012indoor, barron2016fast, zhang2018deepdepth} on LiDAR with different sparsity are shown in \figref{fig:sparsity}. We can see the performances are better (i.e. lower RMSE) with more input sparse depth. Our method still performs reasonably well even for the most challenging case (i.e. 0.0168\%). Actually our result under this case is still better than results of the traditional methods \cite{silberman2012indoor, barron2016fast} with full sparse data (i.e. 4.3\%).


\vspace{2.0pt}
\noindent\textbf{Depth Completion in Indoor Scenes.}
We also evaluate our model for indoor scenes on NYUv2 dataset \cite{silberman2012indoor}.
Adopting similar experiment setting as \cite{cheng2018depth,Ma2018SparseToDense}, we synthetically generate sparse depth via random sampling, train on 50K images sampled from the training set, and evaluate on the official labeled test set (containing 654 images).
Images are down-sampled to half resolution and center-cropped to 304 $\times$ 228. 
The same metrics are adopted, including root mean square error (RMSE), mean absolute relative error (REL), and the percentage $\delta_t$ of pixels with both the relative 
error and inverse of it under a certain threshold $t$ (where $t=1.25, 1.25^2, 1.25^3$).
The quantitative comparisons are listed in \tabref{tab:nyu}.
The numbers for Bilateral \cite{silberman2012indoor}, Ma \etal \cite{Ma2018SparseToDense}, and CSPN \cite{cheng2018depth} are obtained from CSPN \cite{cheng2018depth}.
The numbers for the other methods are obtained using their released implementations.
Even not designed specifically for indoor environment, our method still achieve comparable or better performance compared to the state-of-the-art (rank top for 4 out of 5 metrics).
Please refer to supplementary materials for more qualitative results.

{
	\setlength{\tabcolsep}{0.3em}
	\begin{table}[t]
	\vspace{-2mm}
		\centering \footnotesize
		\begin{tabular}{l|c|c|c|c|c}
			\hline
			& \textbf{RMSE}$\downarrow$ & REL$\downarrow$ & $\delta_{1.25}\uparrow$  & $\delta_{1.25^2}\uparrow$  & $\delta_{1.25^3}\uparrow$    \\ \cline{1-4}
			\hline
			Bilateral \cite{silberman2012indoor} &0.479 & 0.084 & 92.4 & 97.6 & 98.9\\
			TGV \cite{ferstl2013image} & 0.635 & 0.123 & 81.9 & 93.0& 96.8\\
			Ma \etal \cite{Ma2018SparseToDense} & 0.230 & 0.044 & 97.1 & 99.4& 99.8\\
			Zhang \etal~\cite{zhang2018deepdepth} & 0.228  & 0.042 & 97.1  & 99.3  & 99.7 \\
			CSPN ~\cite{cheng2018depth} & 0.117  & \textbf{0.016} & 99.2  &  \textbf{99.9}  &  \textbf{100} \\
			\hline
			Ours &  \textbf{0.115} &  0.022 & \textbf{99.3} & \textbf{99.9}  & \textbf{100} \\
			\hline
		\end{tabular}
		\vspace{0.5em}
		\caption{{\bf Performance of depth completion on NYU v2 dataset \cite{silberman2012indoor}.} We compare to non-learning based approaches \cite{silberman2012indoor,ferstl2013image} and deep learning based methods \cite{Ma2018SparseToDense, cheng2018depth, zhang2018deepdepth}. 
		Our method performs the best on the main RMSE metric (in meter), and performs on-par with the state-of-art method on other metrics. }
		\label{tab:nyu}
	\end{table}
	
}

\vspace{-0.2em}
\section{Conclusion}
\vspace{-0.2em}
In this paper, we propose an end-to-end neural network for depth prediction from Sparse LiDAR data and a single color image. 
We use the surface normal as the intermediate representation directly in the network and demonstrate it is still effective for outdoor scene similar as the indoor scene. 
We propose a deep completion unit to better fuse the color image with the sparse input depth.
We also analyze specific challenges for the outdoor scene, and provide solutions within the network architecture, such as attention based integration to improve performance in far distance and estimating a confidence mask for occlusion handling.
Extensive experiments show that our method achieves the state-of-art performance on the benchmark, and generalizes well to sparser input and indoor scenes.
%

{
\noindent\textbf{Acknowledgements:}
This work was supported in part by National Foundation of China under Grants 61872067 and 61720106004, in part by Department of Science and Technology of Sichuan Province under Grant 2019YFH0016.
}

{\small
	\bibliographystyle{ieee}
	\bibliography{bibliography_long.bib,egbib,egbib_yinda}
}

\end{document}